\documentclass{article}
\PassOptionsToPackage{sort&compress}{natbib}

\newcommand{\KVSet}{\mathbf{KV}}
\newcommand{\ModelOuput}{\mathbf{X}_{\theta}}
\newcommand{\KVOutput}{\mathbf{kv}}

\usepackage{graphicx}
\usepackage{amsmath, amssymb}
\usepackage{algorithm}
\usepackage{algpseudocode}
\usepackage{multirow}
\usepackage[table]{xcolor}
\usepackage{tablefootnote}
\usepackage{subcaption}

\usepackage{wrapfig}
\usepackage{lipsum} 
\usepackage{booktabs} 
\usepackage{adjustbox}

\definecolor{catgray}{gray}{0.92}
\usepackage[final]{neurips_2025}

\usepackage[utf8]{inputenc} 
\usepackage[T1]{fontenc}    
\usepackage{hyperref}       
\usepackage{url}            
\usepackage{booktabs}       
\usepackage{amsfonts}       
\usepackage{nicefrac}       
\usepackage{microtype}      
\usepackage{xcolor}         
\usepackage{multirow}

\title{Self Forcing: Bridging the Train-Test Gap in \\ Autoregressive Video Diffusion}
\author{%
Xun Huang$^1$ \quad Zhengqi Li$^1$ \quad Guande He$^2$ \quad Mingyuan Zhou$^2$ \quad Eli Shechtman$^1$ \\
$^1$Adobe Research \quad $^2$The University of Texas at Austin\\\\
\centerline{\url{https://self-forcing.github.io/}}
}

\begin{document}

\maketitle

\vspace{-1em}
\begin{abstract}
  We introduce Self Forcing, a novel training paradigm for autoregressive video diffusion models. It addresses the longstanding issue of exposure bias, where models trained on ground-truth context must generate sequences conditioned on their own imperfect outputs during inference. Unlike prior methods that denoise future frames based on ground-truth context frames, Self Forcing conditions each frame’s generation on previously self-generated outputs by performing autoregressive rollout with key-value (KV) caching during training. This strategy enables supervision through a holistic loss at the video level that directly evaluates the quality of the entire generated sequence, rather than relying solely on traditional frame-wise objectives. To ensure training efficiency, we employ a few-step diffusion model along with a stochastic gradient truncation strategy, effectively balancing computational cost and performance. We further introduce a rolling KV cache mechanism that enables efficient autoregressive video extrapolation. Extensive experiments demonstrate that our approach achieves real-time streaming video generation with sub-second latency on a single GPU, while matching or even surpassing the generation quality of significantly slower and non-causal diffusion models.  
\end{abstract}

\section{Introduction}
Recent years have witnessed tremendous progress in video synthesis, with state-of-the-art systems now capable of generating remarkably realistic content with complex temporal dynamics~\cite{videoworldsimulators2024}.
However, these results are typically achieved with diffusion transformers~(DiT)~\cite{peebles2022scalable,wang2025wan} that denoise all frames simultaneously using bidirectional attention. This design allows the future to affect the past and requires generating the entire video at once, fundamentally limiting their applicability to real-time streaming applications where future information is unknown when generating the current frame.

In contrast, autoregressive~(AR) models~\cite{yan2021videogpt,ge2022long,yu2023language,hong2022cogvideo, kondratyuk2024videopoet} generate videos sequentially, a paradigm that naturally aligns with the causal structure of temporal media.
This approach not only significantly reduces the viewing latency of generated videos but also unlocks numerous applications, including real-time interactive content creation~\cite{chen2024streaming,liang2025looking}, game simulation~\cite{oasis2024,yu2025gamefactory,parkerholder2024genie2, valevski2025diffusion}, and robotics learning~\cite{yang2023learning, yu2024learning, li2025unified}.
However, AR models often struggle to match the visual fidelity achieved by state-of-the-art video diffusion models due to their reliance on lossy vector quantization techniques~\cite{van2017neural}.

\begin{figure*}
  \centering
  \includegraphics[trim=55pt 0pt 105pt 0pt,width=\textwidth]{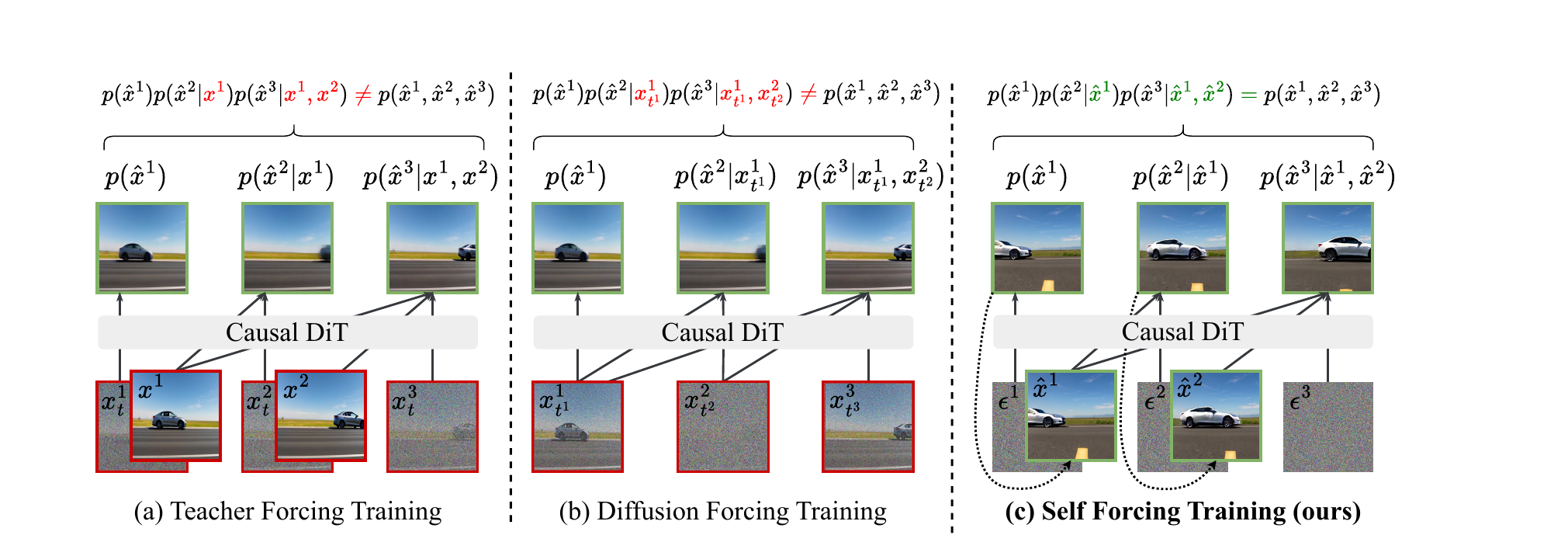}
  \caption{
    \textbf{Training paradigms for AR video diffusion models.}
    (a) In Teacher Forcing, the model is trained to denoise each frame conditioned on the preceding clean, ground-truth context frames.
    (b) In Diffusion Forcing, the model is trained to denoise each frame conditioned on the preceding context frames with varying noise levels.
    Both (a) and (b) generate outputs that do not belong to the distribution the model generates during inference.
    (c) Our Self Forcing approach performs autoregressive self-rollout \textit{during training}, denoising the next frame based on previous context frames generated by itself. A distribution-matching loss~(\textit{e.g.}, SiD, DMD, GAN) is computed on the final output video to align the distribution of generated videos with that of real videos.
    Our training paradigm closely mirrors the inference process, thereby bridging the train-test distribution gap.
  }
  \vspace{-1.5em}
  \label{fig:main}
\end{figure*}

To combine the best of both worlds, two recent techniques have emerged to equip video diffusion models with AR generation capabilities: Teacher Forcing (TF)~\cite{hu2024acdit,jin2024pyramidal,gao2024ca2,zhang2025test} and Diffusion Forcing (DF)~\cite{chen2024diffusion,yin2025causvid,song2025history,gu2025long,magi1,chen2025skyreels}.
Teacher Forcing, a well-established paradigm in sequence modeling, trains the model to predict the next token conditioned on ground-truth tokens.
When applied to video diffusion, TF involves denoising each frame using clean, ground-truth context frames (Figure~\ref{fig:main} (a)), a strategy commonly referred to as next-frame prediction.
In contrast, Diffusion Forcing trains the model on videos with noise levels independently sampled for each frame, denoising each frame based on noisy context frames~(Figure~\ref{fig:main}~(b)).
This ensures the autoregressive inference scenario, where context frames are clean and the current frame is noisy, is covered by the training distribution.

However, models trained with TF or DF often suffer from error accumulation during autoregressive generation,
leading to degraded video quality over time~\cite{yin2025causvid,zhang2025packing,wang2025error}. This issue is more broadly known as \textit{exposure bias}~\cite{schmidt2019generalization,ning2024elucidating}, where a model is trained exclusively on ground-truth context but must rely on its own imperfect predictions at inference time, resulting in a distributional mismatch that compounds errors as generation progresses. While some approaches attempt to mitigate this issue in video diffusion models by incorporating noisy context frames during inference~\cite{chen2024diffusion,oasis2024,zhang2025packing},
such design sacrifices temporal consistency, complicates the KV-cache design, increases generation latency, and does not fundamentally resolve the exposure bias problem.

In this work, we propose \textit{Self Forcing}~(SF), a novel algorithm addressing exposure bias in autoregressive video generation.
Inspired by early RNN-era sequence modeling techniques~\cite{lamb2016professor,ranzato2016sequence,yu2017seqgan}, our approach bridges the train-test distribution gap by explicitly unrolling autoregressive generation during training, generating each frame conditioned on previously self-generated frames rather than ground-truth ones.
This enables supervision with holistic distribution-matching losses~\cite{yin2024one,yin2024improved,goodfellow2014generative} applied to complete generated video sequences.
By forcing the model to encounter and learn from its own prediction errors, Self Forcing effectively mitigates exposure bias and reduces \mbox{error accumulation.}

While Self Forcing may seem computationally prohibitive due to its sequential nature preventing parallel training, we demonstrate that it can be efficiently implemented as an algorithm in the post-training stage where the model does not require a large number of gradient updates to converge.
By employing a few-step diffusion backbone and a carefully designed gradient truncation strategy, Self Forcing is surprisingly more efficient than alternative parallel strategies, achieving superior performance within the same wall-clock training time.
Additionally, we introduce a rolling KV cache mechanism that enhances the efficiency of video extrapolation.

Extensive experiments demonstrate that our model enables real-time video generation at 17 FPS with sub-second latency on a single H100 GPU, while achieving competitive or superior generation quality compared to recent slow bidirectional and autoregressive video diffusion models.
These advances open the door to genuinely interactive video generation use cases—live streaming, gaming, and world simulation—where latency budgets are measured in milliseconds rather than minutes.

\section{Related Work}
\noindent\textbf{GANs for Video Generation.}
Early video generation approaches relied primarily on generative adversarial networks~(GANs)~\cite{goodfellow2014generative}, either using convolutional networks to generate entire videos in parallel~\cite{vondrick2016generating,saito2017temporal,brooks2022generating} or employing recurrent architectures to produce frames sequentially~\cite{villegas2017decomposing,denton2017unsupervised,tulyakov2018mocogan,liu2021infinite, li2022infinitenature}.
Recently, GANs have also been applied to distill video diffusion models~\cite{zhang2024sf,lin2025diffusion,mao2025osv,wu2025snapgen}.
Since the generator in GANs follows the same process during training and inference, it inherently avoids exposure bias. Our work draws inspiration from this fundamental GAN principle by directly optimizing the alignment between the generator's output distribution and the target distribution.

\noindent\textbf{Autoregressive/Diffusion Models for Video Generation.}
Modern video generation models have largely shifted toward diffusion or autoregressive models due to their stronger scaling abilities. Video diffusion models typically adopt bidirectional attention mechanisms to simultaneously denoise all video frames~\cite{ho2022video,blattmann2023align,Ho2022ImagenVH,videoworldsimulators2024,polyak2024movie,yang2025cogvideox,hacohen2024ltx,kong2024hunyuanvideo,blattmann2023stable,villegas2023phenaki,deng2024nova,gupta2024photorealistic,wang2025wan}. Autoregressive models, in contrast, are trained with next-token prediction objectives and generate spatiotemporal tokens sequentially at inference time~\cite{weissenborn2020scaling, kondratyuk2024videopoet,yan2021videogpt,wang2024loong,Bruce2024GenieGI,ren2025next}.

\noindent\textbf{Autoregressive-Diffusion Hybrid Models.}
Very recently, hybrid models integrating autoregressive and diffusion frameworks have emerged as a promising direction in generative modeling of videos~\cite{weng2024art,liu2024redefining,chen2024diffusion,guo2025long,yin2025causvid,hu2024acdit,jin2024pyramidal,gu2025long,gao2024ca2,li2025arlon,liu2024mardini,zhang2025generative,zhang2025test} as well as other sequence domains \cite{li2024autoregressive,wu2023ar,deng2024causal,arriola2025block,liu2024autoregressive,zhou2025transfusion,mo2025xfusionintroducingnewmodality}.
They typically rely on a long, iterative prediction chain~(both temporally autoregressive and spatially iterative denoising), which could lead to significant error accumulation.
Our work addresses this issue
by training the model conditioned on its own predictions and teaching it to correct its own mistakes.

\noindent\textbf{Rolling Diffusion and Variants.}
Another line of work~\cite{ruhe2024rolling,kim2025fifo,xie2024progressive,zhang2025packing,magi1,sun2025ar} trains video diffusion models with a progressive noise schedule, where the noise level gradually increases from earlier to later frames. While these methods support sequential long video generation with less accumulated errors and are sometimes also referred to as autoregressive, they do not strictly follow the autoregressive chain rule decomposition. Consequently, they would exhibit significant latency in interactive applications, as future frames are partially pre-generated before the current frame is presented to the user. This premature commitment restricts the impact of real-time user-injected controls, resulting in limited responsiveness in immediately subsequent frames.

\noindent\textbf{CausVid.}
Our work is most closely related to CausVid~\cite{yin2025causvid}, which trains few-step autoregressive diffusion models using the DF scheme and distribution matching distillation (DMD).  However, CausVid suffers from a critical flaw that its training outputs (generated via DF) do not come from the distribution the model produces at inference time, therefore the DMD loss is matching the wrong distribution. We pinpoint this issue and propose a solution that matches the true model distribution.

\section{Self Forcing: Briding Train-Test Gap via Holistic Post-Training}
We first provide a formal definition of autoregressive video diffusion models and describe standard training approaches in Section~\ref{sec:preliminaries}. In Section~\ref{sec:rollout}, we introduce the main part of our Self Forcing training algorithm and describe how it can be efficiently implemented with a few-step diffusion model.
In Section~\ref{sec:loss}, we describe various choices of holistic, video-level distribution-matching training objectives. Finally, we introduce a rolling key-value cache mechanism that enables efficient generation of arbitrarily long videos in Section~\ref{sec:rolling_kv}.

\subsection{Preliminaries: Autoregressive Video Diffusion Models}
\label{sec:preliminaries}
Autoregressive video diffusion model is a hybrid generative model that combines autoregressive chain-rule decomposition with denoising diffusion models for video generation. 
Specifically, given a sequence of $N$ video frames \( x^{1:N} = (x^1, x^2, \dots, x^N) \), it factorizes the joint distribution into product of conditionals using the chain rule $p(x^{1:N}) = \prod_{i=1}^{N} p(x^i|x^{<i})$.
Each conditional distribution \( p(x^i|x^{<i}) \) is then modeled using a diffusion process, where a frame is generated by progressively denoising an initial Gaussian noise conditioned on previously generated frames.
This formulation combines the strengths of both autoregressive models and diffusion models for capturing sequential dependencies while enabling high-quality generation of continuous-valued visual signal.
In practice, we can also choose to generate one chunk of frames rather than a single frame at a time~\cite{yin2025causvid,magi1}. For simplicity of notation, however, we continue to denote each chunk as a frame in this section.

Most existing autoregressive video diffusion models are trained using frame-wise denoising loss within the paradigm of Teacher Forcing (TF) or Diffusion Forcing (DF).
Specifically, each frame \(x^i\) is corrupted by the forward process \( q_{t^i|0}(x_{t^i}^{i} | x_0^{i}) \) such that \(x_{t^i}^i  = \Psi(x^i, \epsilon^i, t^i) = \alpha_{t^i} x^i + \sigma_{t^i} \epsilon^i \), where \( \alpha_{t^i}, \sigma_{t^i} \) are pre-defined noise schedule within a finite time horizon \( t^i \in [0, 1000] \) and \( \epsilon^i \sim \mathcal{N}(0, I) \) is Gaussian noise. In TF, the timesteps $t^i$ are usually shared across all frames, whereas in DF, they are sampled independently for each frame.
A generative model is learned through the time-reversal of the forward process, where each denoising step can be achieved by predicting the noise $\epsilon^i$ added to each frame with a neural network \(\hat \epsilon_\theta^i := G_\theta(x_{t^i}^i, t^i, c) \) conditioned on the context $c$.
The context consists of clean ground-truth frames \( x^{<i} \) in TF or noisy context frames \( x^{j<i}_{t^j} \) in DF.
The model is trained to minimize the frame-wise mean squared error (MSE) between the predicted noise and the true added noise:
\( \mathcal{L}^{\text{DM}}_\theta = \mathbb{E}_{x^i, t^i, \epsilon^i} \left[ w_{t^i} \| \hat{\epsilon}^i_\theta - \epsilon^i \|_2^2 \right] \), where $w_{t^i}$ is a weighting function.

\begin{figure*}
  \centering
  \includegraphics[trim=35pt 5pt 65pt 5pt,width=\textwidth]{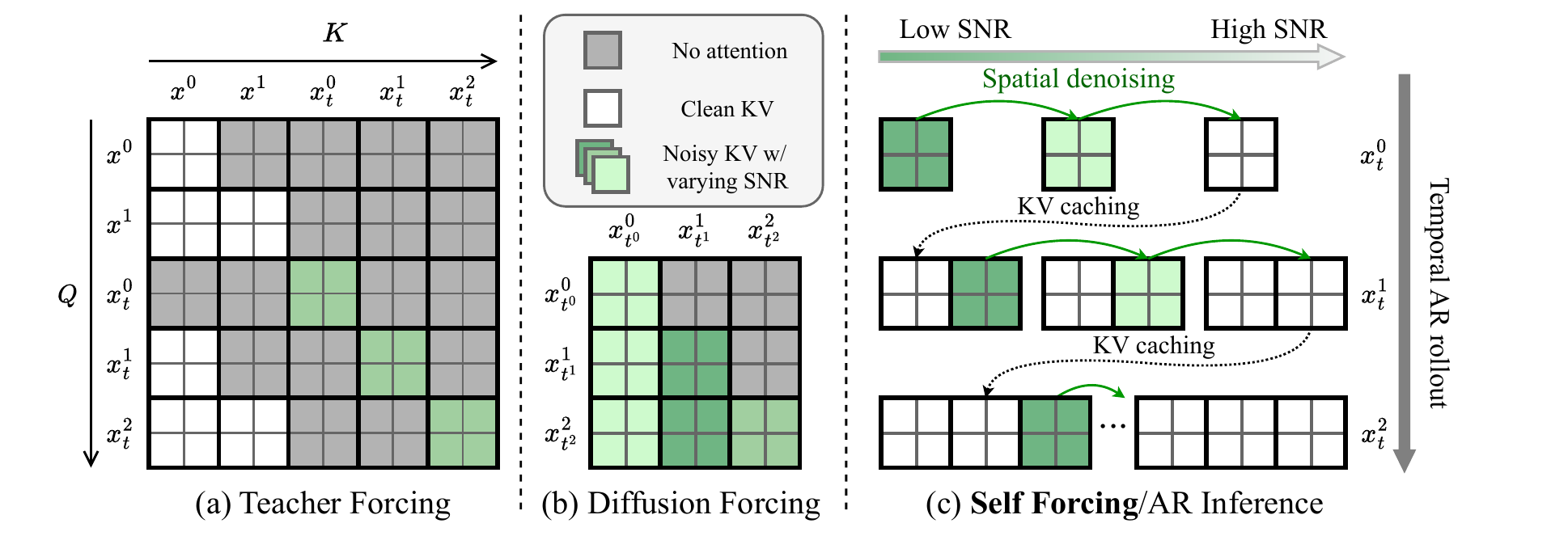}
  \caption{
    \textbf{Attention mask configurations.}  Both Teacher Forcing (a) and Diffusion Forcing (b) train the model on the entire video in parallel, enforcing causal dependencies with custom attention masks. In contrast, our Self-Forcing Training (c) mirrors the autoregressive (AR) inference process with KV caching and does not rely on special attention masks. For illustration purposes, we show a scenario where the video contains 3 frames, and each frame consists of 2 tokens.
  }
  \vspace{-1em}
  \label{fig:attention}
\end{figure*}

We focus on the transformer-based architecture~\cite{peebles2022scalable} of diffusion models with text conditioning~(omitted from equations for clarity) operating in a compressed latent space encoded by a causal 3D variational autoencoder (VAE)~\cite{kingma2013auto}. The autoregressive chain-rule decomposition is implemented via causal attention. Figures~\ref{fig:attention} (a) and (b) illustrate the attention mask configurations of Teacher Forcing and Diffusion Forcing approaches. For Teacher Forcing, we describe an efficient variant that processes all frames in parallel using block sparse attention masks, rather than denoising one frame at each training iteration~\cite{jin2024pyramidal}. Such design has been used in MAR-based~\cite{li2024autoregressive} autoregressive video generation~\cite{zhou2025taming} and concurrently in other autoregressive video diffusion models~\cite{zhang2025generative,zhang2025test}.

\subsection{Autoregressive Diffusion Post-Training via Self-Rollout}
\label{sec:rollout}
The core idea of Self Forcing is to generate videos through autoregressive self-rollout during training following the inference-time recipe. Specifically, we sample a batch of videos $\{x^{1:N}_{\theta}\}\sim p_\theta(x^{1:N}) = \prod_{i=1}^{N} p_\theta(x^i|x^{<i})$ where each frame \( x^i \) is generated by performing iterative denoising conditioned on self-generated outputs, including both clean context frames in the past and noisy frames at the current time step.
Unlike most previous autoregressive models that only utilize KV caching during inference, our Self Forcing method innovatively employs KV caching during training, as shown in Figure~\ref{fig:attention}~(c).

Nevertheless, implementing Self Forcing with standard many-step diffusion models would be computationally prohibitive, as it requires unrolling and backpropagation through long denoising chains.
Therefore, we choose to use a few-step diffusion model $G_\theta$ to approximate each conditional distribution $p_\theta(x^i|x^{<i})$ in the autoregressive factorization. Consider $\{t_0=0,t_1, \dots, t_{T}=1000\}$ a subsequence of timesteps $[0, ..., 1000]$,
at each denoising step $t_j$ and frame index $i$, the model denoises an intermediate noisy frame \( x^i_{t_j} \) conditioned on previous clean frames \( x^{<i} \). It then injects Gaussian noise with a lower noise level into the denoised frame through the forward process $\Psi$ to obtain the noisy frame \( x^i_{t_{j - 1}} \) as the input to the next denoising step, following the standard practice in few-step diffusion models~\cite{song2023consistency,yin2024improved}.
The model distribution $p_\theta(x^i|x^{<i})$ is implicitly defined as $ f_{\theta,t_1} \circ f_{\theta,t_2} \circ ... \circ f_{\theta,t_T} (x^i_{t_T})$, where $f_{\theta,t_j}(x_{t_j}^i)=\Psi(G_\theta(x^i_{t_j}, t_{j},  x^{<i}), \epsilon_{t_{j-1}}, t_{j-1})$, \mbox{and $ x^i_{t_{T}} \sim \mathcal{N}(0, I)$}.
\begin{figure}[t!]
\begin{minipage}[t]{0.48\textwidth}
  \begin{algorithm}[H]
    \caption{Self Forcing Training}
    \small
    \begin{algorithmic}[1]
      \Require Denoise timesteps $\{t_1, \dots, t_T\}$
      \Require Number of video frames $N$
      \Require AR diffusion model $G_\theta$ (returns KV embeddings via $G_\theta^{\mathrm{KV}}$)
      \Loop
        \State Initialize model output $\ModelOuput \gets []$
        \State Initialize KV cache $\KVSet \gets []$
        \State Sample $s \sim \text{Uniform}(1, 2, \ldots, T)$
        \For{$i = 1, \dots, N$}
          \State Initialize $x^i_{t_T} \sim \mathcal{N}(0, I)$
          \For{$j = T, \dots, s$}
            \If{$j = s$}
              \State Enable gradient computation
              \State Set $\hat{x}^i_{0} \gets G_\theta(x^i_{t_{j}}; t_j, \KVSet)$
              \State $\ModelOuput{\texttt{.append}}(\hat x^i_{0})$
              \State Disable gradient computation
              \State Cache $\KVOutput^i \gets G_\theta^\text{KV}(\hat{x}^i_{0}; 0, \KVSet)$
              \State $\KVSet{\texttt{.append}}(\KVOutput^i)$
            \Else
              \State Disable gradient computation
              \State Set $\hat{x}^i_{0} \gets G_\theta(x^i_{t_j}; t_j, \KVSet)$
              \State Sample $\epsilon \sim \mathcal{N}(0, I)$
              \State Set $x^i_{t_{j-1}} \gets \Psi(\hat{x}^i_0, \epsilon, t_{j-1})$
            \EndIf
          \EndFor
        \EndFor
        \State Update $\theta$ via distribution matching loss
      \EndLoop
    \end{algorithmic}
    \label{alg:self_forcing}
  \end{algorithm}
\end{minipage}
\hfill
\begin{minipage}[t]{0.48\textwidth}
  \begin{algorithm}[H]
    \caption{Autoregressive Diffusion Inference with Rolling KV Cache}
    \small
    \begin{algorithmic}[1]
      \Require KV cache of size $L$ frames
      \Require Denoise timesteps $\{t_1, \dots, t_T\}$
      \Require Number of generated frames $M$
      \Require AR diffusion model $G_\theta$ (returns KV embeddings via $G_\theta^\text{KV}$)
      \State Initialize model output $\ModelOuput \gets []$
      \State Initialize KV cache $\KVSet \gets []$
      \For{$i = 1, \dots, M$}
        \State Initialize $x^i_{t_T} \sim \mathcal{N}(0, I)$
        \For{$j = T, \dots, 1$}
          \State Set $\hat{x}^i_{0} \gets G_\theta(x^i_{t_j}; t_j, \KVSet)$
          \If{$j = 1$}
            \State $\ModelOuput{\texttt{.append}}(\hat x^i_{0})$
            \State Cache $\KVOutput^i \gets G_\theta^\text{KV}(\hat{x}^i_{0}; 0, \KVSet)$
            \If{$|\KVSet| = L$}
              \State $\KVSet.\mathrm{pop}(0)$ \Comment{Cache eviction}
            \EndIf
            \State $\KVSet{\texttt{.append}}(\KVOutput^i)$
          \Else
            \State Sample $\epsilon \sim \mathcal{N}(0, I)$
            \State Set $x^i_{t_{j-1}} \gets \Psi(\hat{x}^i_0, \epsilon, t_{j-1})$
          \EndIf
        \EndFor
      \EndFor
      \State \Return $\ModelOuput$
    \end{algorithmic}
    \label{alg:inference}
  \end{algorithm}
\end{minipage}
\vspace{-1em}
\end{figure}

Even with few-step models, however, naively backpropagating through the entire autoregressive diffusion process would still lead to excessive memory consumption.
To address this challenge, we propose a gradient truncation strategy that limits the backpropagation to only the final denoising step of each frame. Moreover, instead of always using $T$ denoising steps~(as in inference time), we randomly sample a denoising step $s$ from $[1, T]$ for each sample sequence at each training iteration, and use the denoised output of the $s$-th step as the final output.
This stochastic sampling approach ensures all intermediate denoising steps receive supervision signals.
We additionally detach the gradients of the previous frames from the current frame during training by restricting gradient flow into KV cache embeddings. For a complete description of the training process, see Algorithm~\ref{alg:self_forcing}.

\subsection{Holistic Distribution Matching Loss}
\label{sec:loss}
Autoregressive self-rollout generates samples directly from the inference-time model distribution, enabling us to apply holistic, video-level losses that align the distribution of generated videos \( p_\theta(x^{1:N}) \) with that of real videos \( p_{\text{data}}(x^{1:N}) \).
To leverage pre-trained diffusion models and enhance training stability~\cite{jenni2019stabilizing}, we inject noise to both distributions and match \( p_{\theta, t}(x^{1:N}_t) \) and \( p_{\text{data}, t}(x^{1:N}_t) \), where each represents the respective distribution after applying the forward diffusion process: \( p_{\cdot, t}(x^{1:N}_t) = \int q_{t|0}(x^{1:N}_t | x^{1:N}) p_{\cdot}(x^{1:N}) \mathrm{d} x^{1:N} \).
Our framework is generally applicable to various divergence measures and distribution matching frameworks, and we consider three approaches in this paper:
\begin{itemize}

  \item \textbf{Distribution Matching Distillation (DMD)}~\cite{yin2024one,yin2024improved}: This approach minimizes the reverse Kullback-Leibler divergence $\mathbb{E}_{t} [ D_\text{KL}(p_{\theta, t} \| p_{\text{data}, t}) ]$ by leveraging the score difference between distributions to guide gradient updates.
  \item \textbf{Score Identity Distillation (SiD)}~\cite{zhou2024score,zhou2025adversarial}: This method performs distribution matching via Fisher divergence $\mathbb E_{t, p_{\theta, t}}[\| \nabla \log p_{\theta, t}\ - \nabla \log p_{\text{data}, t}\|^2]$.

  \item \textbf{Generative Adversarial Networks (GANs)}~\cite{goodfellow2014generative}: It approximately minimizes the Jensen-Shannon divergence through a minimax game between the generator (our autoregressive diffusion model) and a discriminator that distinguishes between real and generated videos.
\end{itemize}

Importantly, our training objective matches the \textit{holistic} distribution of the entire video sequence to the data distribution $D(p_{\text{data}}(x^{1:N}) \| p_\theta(x^{1:N}))$. In contrast, TF/DF can be understood as performing \textit{frame-wise} distribution matching: $\mathbb{E}_{\{x^{<i}\}\sim p_{\text{data}}} D_{KL}(p_{\text{data}}(x^{i}|x^{<i}) \| p_\theta(x^{i}|x^{<i}))$\footnote{With a specific weighting per noise level~\cite{kingma2021variational,song2021maximum}, denoising loss approximates the maximum likelihood objective, equivalent to minimizing the KL divergence between per-frame data and model distributions.}, where DF additionally samples context frames from a noise-corrupted data distribution $\{x^{<i}\}\sim\tilde{p}_{\text{data}}$.
Our formulation fundamentally transforms the training dynamics—context frames $\{x^{<i}\}$ are sampled from the model's own distribution $p_\theta$ rather than from the data distribution (clean or noisy). This alignment between training and inference distributions effectively addresses exposure bias and forces the model to learn from its own imperfections, thereby developing robustness to error accumulation.

While all three objectives have been used in the context of timestep distillation of diffusion models, our primary motivation differs fundamentally from distillation: we aim to enhance the \textit{quality} of autoregressive video generation by addressing exposure bias via distribution matching, rather than merely accelerating sampling.
This distinction makes other popular distillation methods~\cite{song2023consistency} inapplicable to our framework as they only focus on timestep reduction without directly aligning the generator output distribution.
Although CausVid~\cite{yin2025causvid} similarly employs DMD to match the distribution of generated videos, the distribution it optimizes during training (using Diffusion Forcing outputs) deviates from the actual inference-time distribution, significantly undermining its effectiveness.

\subsection{Long Video Generation with Rolling KV Cache}
\label{sec:rolling_kv}

\begin{figure*}[t]
  \centering
  \includegraphics[trim=15pt 10pt 15pt 10pt,width=\textwidth]{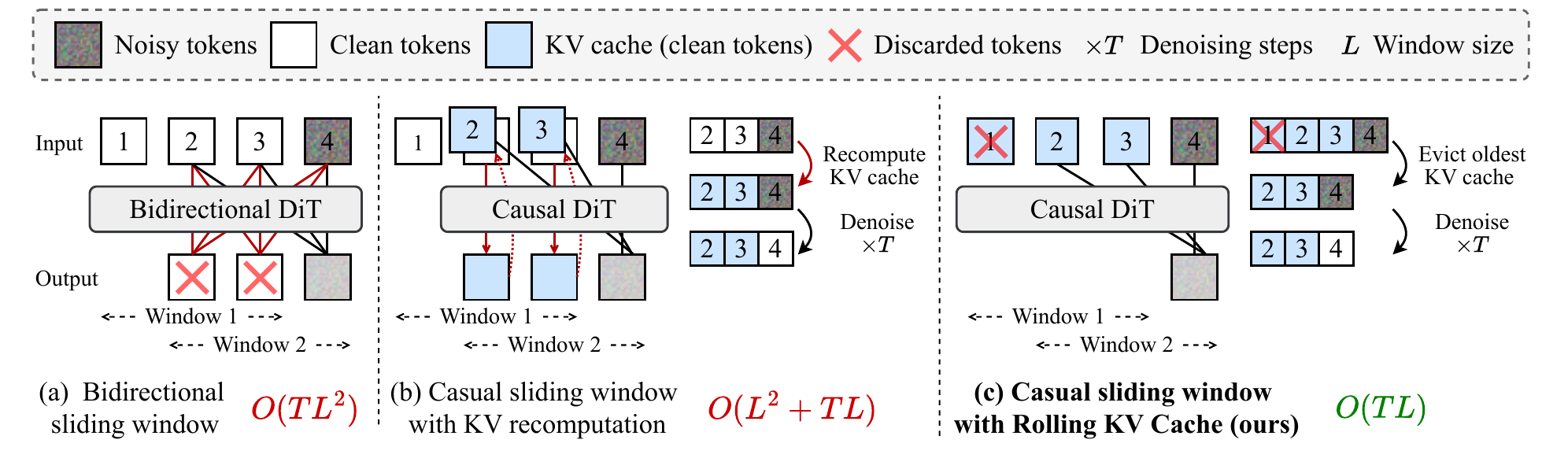}
  \caption{
    \textbf{Efficiency comparisons for video extrapolation.}
    When performing video extrapolation through sliding window inference, (a) bidirectional diffusion models trained with TF/DF~\cite{song2025history,chen2025skyreels} do not support KV cache. (b) Prior causal diffusion models~\cite{yin2025causvid,magi1} require re-computing KV when shifting the window. (c) Our method does not recompute KV and enables \mbox{more efficient extrapolation.}
  }
  \label{fig:kv}
  \vspace{-1em}
\end{figure*}

A key advantage of autoregressive models over standard video diffusion models is their extrapolative ability, in principle allowing the generation of infinitely long videos via sliding-window inference.
While bidirectional attention models trained with Diffusion Forcing~\cite{song2025history,chen2025skyreels} can also generate videos autoregressively, they do not support KV caching, requiring complete recomputation of attention matrices for each new frame. This leads to excessive computational complexity of $O(TL^2)$ (where $T$ represents the number of denoising steps and $L$ is thewindow size), as shown in Figure~\ref{fig:kv}~(a).

Models with causal attention, on the other hand, can leverage KV caching to improve efficiency.
However, existing implementations~\cite{yin2025causvid,magi1} require recomputing KV cache for overlapping frames between consecutive sliding windows, as illustrated in Figure~\ref{fig:kv}~(b).
This leads to $O(L^2+TL)$ complexity when employing dense sliding windows.
As a result, prior implementations adopt larger strides with minimal overlap to reduce computational costs, which compromises temporal consistency since frame at the beginning of each window relies on a significantly reduced historical context.

Inspired by research in large language models~\cite{streamingllm}, we propose a rolling KV cache mechanism for autoregressive diffusion models that allows infinitely long video generation without any need of recomputing the KV cache. As illustrated in Figure~\ref{fig:kv}~(c), we maintain a fixed-size KV cache that stores the KV embeddings of tokens in the most recent $L$ frames. When generating a new frame, we first check if the KV cache is full. If it is, we remove the oldest KV cache entry before adding the new one. This approach enables endless frame generation with a time complexity of $O(TL)$, while still maintaining a sufficient context length when generating each new frame. Algorithm~\ref{alg:inference} provides a detailed description of our autoregressive long video generation algorithm with rolling KV cache.

However, naive implementation of this mechanism leads to severe flickering artifacts due to distribution mismatch. Specifically, the first latent frame has different statistical properties than other frames: it only encodes the first image without performing temporal compression. The model, having always seen the first frame as the image latent during training, fails to generalize when the image latent is no longer visible in the rolling KV cache scenario.
Our solution is straightforward but effective: during training, we restrict the attention window so the model cannot attend to the first chunk when denoising the final chunk, thereby simulating the conditions encountered during long video generation.

\section{Experiments}
\vspace{-0.5em}
\label{sec:experiments}

\noindent\textbf{Implementation.}
We implement Self Forcing with Wan2.1-T2V-1.3B~\cite{wang2025wan}, a Flow Matching~\cite{lipmanflow} based model that generates 5s videos at 16 FPS with a resolution of $832\times480$.
Following CausVid's initialization protocol~\cite{yin2025causvid}, we first finetune the base model with causal attention masking on 16k ODE solution pairs sampled from the base model.
For both ODE initialization and Self Forcing training, we sample text prompts from a filtered and LLM-extended version of VidProM~\cite{wangvidprom}.
We use 4-step diffusion and implement both frame-wise and chunk-wise autoregressive variants, with the latter generating a chunk of 3 latent frames at a time.
We adopt the R3GAN~\cite{huang2024gan} objective, which consists of relativistic pairing GAN loss~\cite{jolicoeurrelativistic} with R1 + R2 regularization~\cite{mescheder2018training}.
We use the 14B base model to generate 70k videos as the dataset for training GANs~\cite{sauer2024fast} and fine-tuning many-step TF/DF AR diffusion baselines. Notably, DMD/SiD implementations of our algorithm remain data-free, capable of converting a pre-trained video diffusion model into an autoregressive model without any video training data.
Additional implementation details are provided in Appendix~\ref{app:implementation}.

\noindent\textbf{Evaluation metrics.}
We adopt VBench~\cite{huang2023vbench} and user preference study to evaluate both visual quality and semantic alignment.
We also rigorously evaluate the efficiency of our method for real-time applications.
While some recent works claim ``real-time'' video generation abilities~\cite{hacohen2024ltx,zhao2025real} based solely on \textit{throughput}, we argue that true real-time performance requires both sufficient throughput (exceeding video playback rate) and lower \textit{latency} than the perceptual threshold which could be application-dependent~\cite{li4654242latency}.
We therefore evaluate both throughput and first-frame latency to provide a comprehensive assessment of real-time capabilities, with all speed tests conducted on a single NVIDIA H100 GPU.

\begin{wrapfigure}{r}{0.48\textwidth}
  \vspace{-1em}
  \centering
  \includegraphics[width=\linewidth]{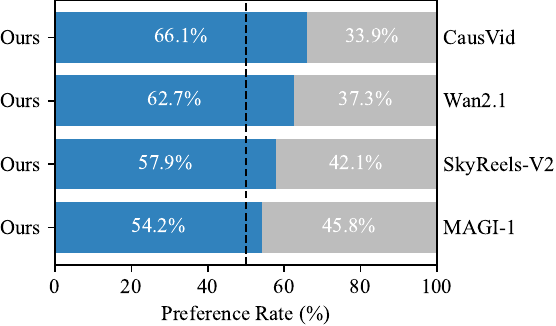}
  \caption{\textbf{User preference study.} Self Forcing outperforms all baselines in human preference.}
  \label{fig:user_study_result}
  \vspace{-1em}
\end{wrapfigure}

\noindent\textbf{Comparison with existing baselines.}
We compare our model with relevant open-source video generation models of similar scale. Our comparisons include two diffusion models: Wan2.1-1.3B~\cite{wang2025wan} (our initialization weights) and LTX-Video~\cite{hacohen2024ltx} (known for efficiency). We also compare with several autoregressive models including Pyramid Flow~\cite{jin2024pyramidal}, NOVA~\cite{deng2024nova}, SkyReels-V2~\cite{chen2025skyreels}, MAGI-1~\cite{magi1}, and CausVid~\cite{yin2025causvid} (also initialized from Wan-1.3B).

\begin{figure}[t!]
  \centering
  \includegraphics[trim=100pt 1pt 1pt 1pt, width=\textwidth]{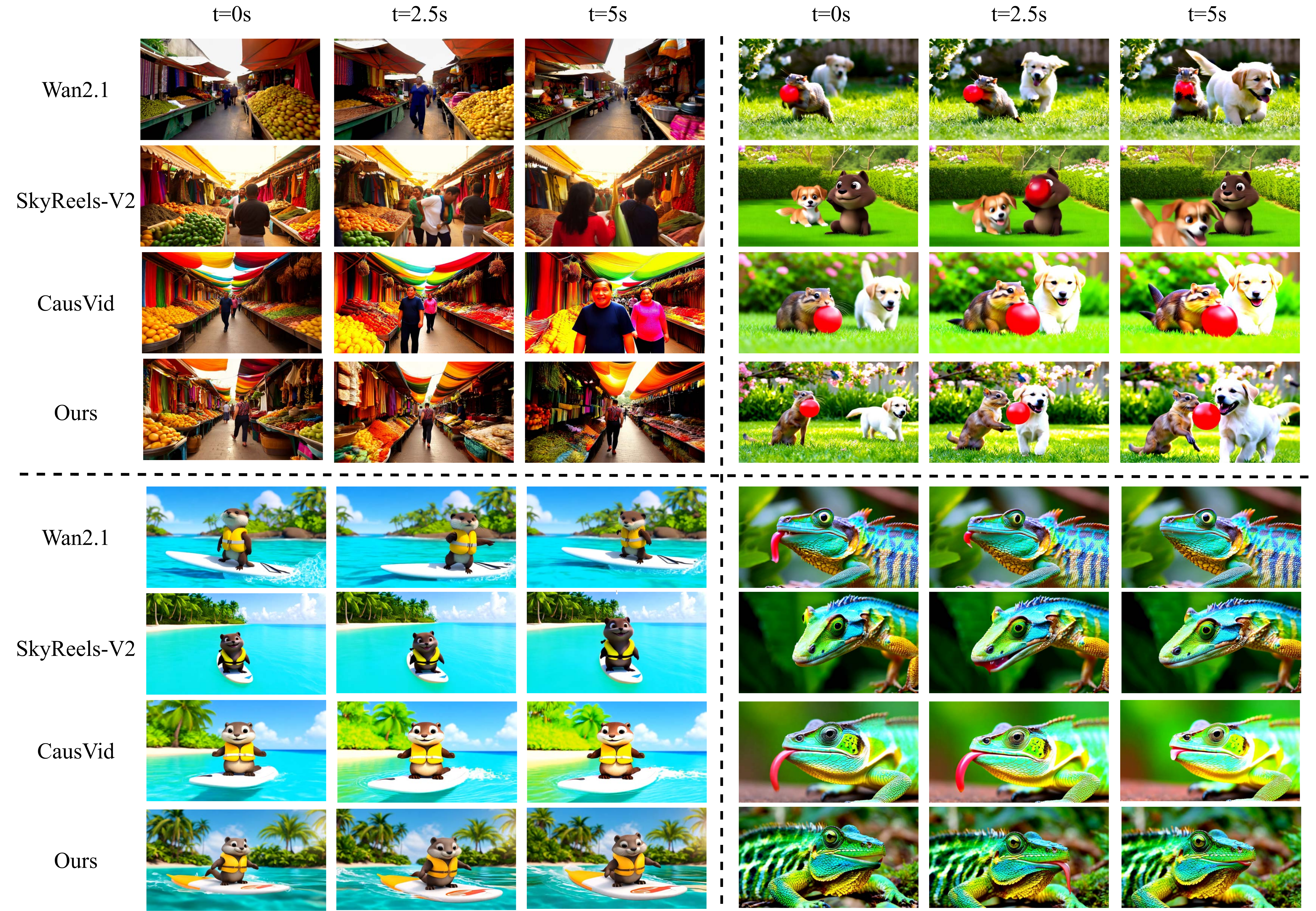}
  \caption{\textbf{Qualitative comparisons.} We visualize videos generated by Self Forcing~(Ours) against those by Wan2.1~\cite{wang2025wan}, SkyReels-V2~\cite{chen2025skyreels}, and CausVid~\cite{yin2025causvid} at three time steps. All models share the same architecture with 1.3B parameters.}
  \label{fig:comp_5s}
  \vspace{-1.5em}
\end{figure}

As shown in Table~\ref{tab:main}, our chunk-wise autoregressive model achieves the highest VBench scores across all compared models while simultaneously delivering real-time throughput ($17.0$ FPS) with sub-second latency, low enough for certain real-time applications such as live video streaming~\cite{bentaleb2025toward}.
Figure~\ref{fig:user_study_result} shows the user study results comparing our chunk-wise Self Forcing model against several important baselines. Our approach is consistently preferred over all alternatives, including the many-step diffusion model Wan2.1 that our model is initialized from.
Our frame-wise variant maintains strong generation quality while providing the lowest latency ($0.45$s), making it particularly suitable for latency-sensitive real-time applications.
Results here are obtained using the DMD loss objective; models trained with SiD and GAN objectives achieve comparable performance as detailed in our ablation studies.
As shown in Figure~\ref{fig:comp_5s}, CausVid suffers from the error accumulation problem that causes the saturation to increase over time. Our approach obtains slightly better visual quality than Wan2.1/SkyReels-V2, while being around \textbf{150x} faster in latency.
More example videos are provided in the project website (\url{https://self-forcing.github.io/}).

\begin{table}[t]
  \small
  \setlength{\tabcolsep}{3.5pt} 
  \caption{
    \textbf{Comparison with relevant baselines.} We compare Self Forcing with representative open-source video generation models of similar parameter sizes and resolutions.
  }
  \vspace{0.5em}
  \label{tab:main}
  \centering
  \begin{tabular}{lcccccccc}
      \toprule
      \multirow{2}{*}{Model} & \multirow{2}{*}{\#Params} & \multirow{2}{*}{Resolution} & \multirow{2}{*}{Throughput} & \multirow{2}{*}{Latency} &
      \multicolumn{3}{c}{Evaluation scores $\uparrow$}\\
    \cmidrule(lr){6-8}
     & & & (FPS) $\uparrow$ & (s) $\downarrow$ & Total & Quality & Semantic \\
     & & &                  &                  & Score & Score & Score \\
    \midrule
    \rowcolor{catgray}
    \multicolumn{8}{l}{\textit{Diffusion models}}\\
    LTX-Video~\cite{hacohen2024ltx}      & 1.9B & $768{\times}512$ & 8.98          & 13.5          & 80.00 & 82.30 & 70.79 \\
    Wan2.1~\cite{wang2025wan}            & 1.3B & $832{\times}480$ & 0.78          & 103           & 84.26 & \textbf{85.30} & 80.09 \\
    \midrule
    \rowcolor{catgray}
    \multicolumn{8}{l}{\textit{Chunk-wise autoregressive models}}\\
    SkyReels-V2~\cite{chen2025skyreels}  & 1.3B & $960{\times}540$ & 0.49          & 112           & 82.67 & 84.70 & 74.53 \\
    MAGI-1~\cite{magi1}                  & 4.5B & $832{\times}480$ & 0.19          & 282           & 79.18 & 82.04 & 67.74 \\
    CausVid~\cite{yin2025causvid}$^{*}$  & 1.3B & $832{\times}480$ & \textbf{17.0} & 0.69          & 81.20 & 84.05 & 69.80 \\
    \midrule
    Self Forcing (Ours, chunk-wise)      & 1.3B & $832{\times}480$ & \textbf{17.0} & 0.69          & \textbf{84.31} & 85.07 & \textbf{81.28}    \\  
    \midrule
    \rowcolor{catgray}
    \multicolumn{8}{l}{\textit{Autoregressive models}$^{\dagger}$}\\
    NOVA~\cite{deng2024nova}             & 0.6B & $768{\times}480$ & 0.88          & 4.1  & 80.12 & 80.39 & 79.05 \\
    Pyramid Flow~\cite{jin2024pyramidal} & 2B   & $640{\times}384$ & 6.7           & 2.5           & 81.72 & 84.74 & 69.62 \\
    \midrule
    Self Forcing (Ours, frame-wise)      & 1.3B & $832{\times}480$ & 8.9           & \textbf{0.45} & 84.26 & 85.25 & 80.30 \\ 
    \bottomrule
  \end{tabular}\\
  \vspace{2pt}
  {\footnotesize\raggedright$^{*}$ We compare with the official implementation of CausVid that uses the same base model~(Wan-1.3B).\par}
  {\footnotesize\raggedright$^{\dagger}$ The distinction of AR/non-AR applies to the temporal dimension.\par}
  \vspace{-1em}
\end{table}

\noindent\textbf{Ablation Studies.}
We perform controlled comparisons  of Self Forcing with alternative autoregressive diffusion training approaches. We evaluate: (1) AR Diffusion models trained with denoising diffusion loss using either Teacher Forcing or Diffusion Forcing, and (2) few-step AR Diffusion models trained with TF/DF inputs but optimized with distribution matching objectives. The latter configuration with DF and DMD essentially replicates CausVid~\cite{yin2025causvid} within our implementation framework, allowing direct comparison under identical training conditions.

Table~\ref{tab:ablation} demonstrates that Self Forcing performs robustly across various distribution matching objectives (DMD, SiD, and GAN), consistently outperforming all baselines.
While baseline methods exhibit notable quality degradation when shifting from chunk-wise to frame-wise AR due to error accumulation associated with increased AR unrolling steps, usually manifesting as progressive over-saturation or over-sharpening (similar to CausVid in Appendix~\ref{app:qualitative} Fig.~\ref{fig:comp_5s}), Self Forcing maintains consistent performance across both setups, highlighting its effectiveness at addressing exposure bias.

\begin{table}[t]
  \caption
  {
    \textbf{Ablation study.} We conduct controlled ablation studies comparing different training paradigms and distribution matching objectives under our training setup across chunk-wise~(left) and frame-wise~(right) AR models.
    Self Forcing works well with all different distribution matching objectives and consistently outperforms alternative training approaches.
  }
  \vspace{0.5em}
  \centering
  \begin{minipage}[t]{0.485\textwidth}
    \centering
    \footnotesize
    \setlength{\tabcolsep}{2pt}
    \begin{tabular}{lccc}
      \toprule
      \multirow{3}{*}{\normalsize{Chunk-wise AR}} &
      \multicolumn{3}{c}{Evaluation scores $\uparrow$}\\
    \cmidrule(lr){2-4}
                           & Total & Quality & Semantic \\
                           & Score & Score & Score \\
    \midrule
    \rowcolor{catgray}
    \multicolumn{4}{l}{\textit{Many~(50$\times$2)-step models}}\\
    Diffusion Forcing~(DF) & 82.95 & 83.66 & 80.09 \\
    Teacher Forcing~(TF)   & 83.58 & 84.34 & 80.52 \\
    \midrule
    \rowcolor{catgray}
    \multicolumn{4}{l}{\textit{Few~(4)-step models}}\\
    DF + DMD  & 82.76 & 83.49 & 79.85 \\
    TF + DMD    & 82.32 & 82.73 & 80.67 \\
    Self Forcing (Ours, DMD) & \textbf{84.31} & 85.07 & \textbf{81.28} \\
    Self Forcing (Ours, SiD) & 84.07 & \textbf{85.52} & 78.24 \\
    Self Forcing (Ours, GAN) & 83.88 & 85.06 & 79.16 \\
    \bottomrule
  \end{tabular}
  \end{minipage}
  \hfill
  \begin{minipage}[t]{0.485\textwidth}
    \centering
    \footnotesize
    \setlength{\tabcolsep}{2pt}
    \begin{tabular}{lccc}
      \toprule
      \multirow{3}{*}{\normalsize{Frame-wise AR}} &
      \multicolumn{3}{c}{Evaluation scores $\uparrow$}\\
    \cmidrule(lr){2-4}
                             & Total & Quality & Semantic \\
                             & Score & Score & Score \\
    \midrule
    \rowcolor{catgray}
    \multicolumn{4}{l}{\textit{Many~(50$\times$2)-step models}}\\
    Diffusion Forcing~(DF)   & 77.24 & 79.72 & 67.33 \\
    Teacher Forcing~(TF)     & 80.34 & 81.34 & 76.34 \\
    \midrule
    \rowcolor{catgray}
    \multicolumn{4}{l}{\textit{Few~(4)-step models}}\\
    DF + DMD                 & 80.56 & 81.02 & 78.71 \\
    TF + DMD                 & 78.12 & 79.62 & 72.11 \\
    Self Forcing (Ours, DMD) & \textbf{84.26} & \textbf{85.25} & \textbf{80.30} \\
    Self Forcing (Ours, SiD) & 83.54          & 84.71          & 78.86          \\
    Self Forcing (Ours, GAN) & 83.27          & 84.57          & 78.08          \\
    \bottomrule
  \end{tabular}
  \end{minipage}
  \label{tab:ablation}
\end{table}




\noindent\textbf{Rolling KV cache.} We observe that recomputing KV cache when shifting sliding window (Fig.~\ref{fig:kv}~(b)) results in significantly reduced throughput (only 4.6 FPS) when generating 10-second videos. While naive rolling KV cache maintains high throughput, it introduces severe visual artifacts, as illustrated in the examples in Appendix~\ref{app:qualitative}. By training the model to generate frames without seeing the initial image latent, we effectively mitigate these artifacts while maintaining high throughput (16.1 FPS).

\begin{figure*}[t]
  \centering
  \includegraphics[width=\textwidth]{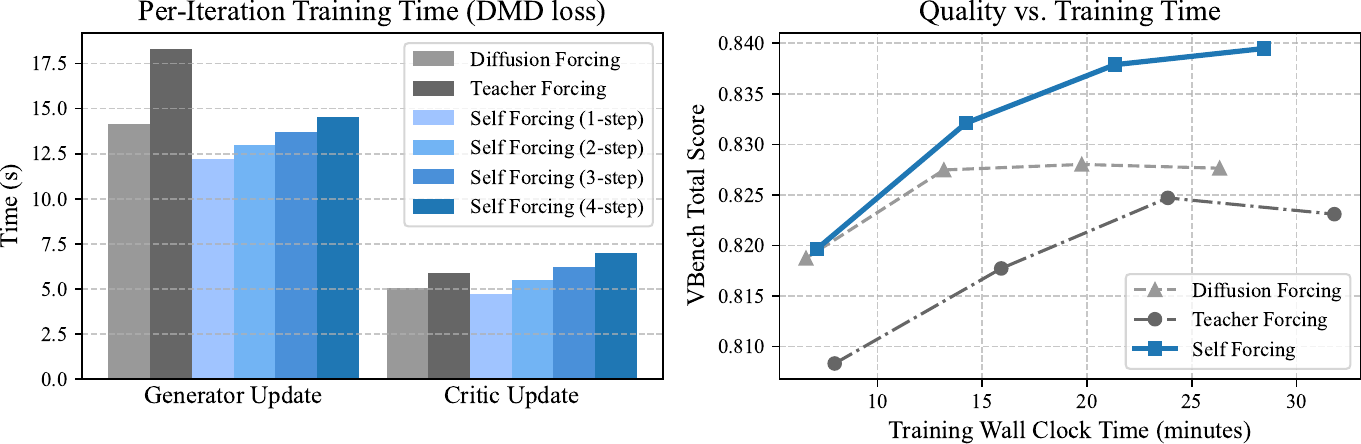}
  \caption{
    \textbf{Training efficiency comparison.} \textit{Left:} Per-iteration time across different chunk-wise, few-step autoregressive video diffusion training algorithms~(using DMD as the distribution matching objective). \textit{Right:} Video quality (VBench score) vs. wall clock training time.
  }
  \label{fig:training_time}
\end{figure*}

\noindent\textbf{Training efficiency.} One might expect Self Forcing training to be computationally prohibitive given its sequential nature that contradicts the parallelizable paradigm of transformers. Surprisingly, our experiments reveal that Self Forcing actually outperforms alternative strategies in training efficiency. As shown in Fig.~\ref{fig:training_time} (left), Self Forcing achieves comparable per-iteration training time to Teacher Forcing and Diffusion Forcing. Furthermore, Fig.~\ref{fig:training_time} (right) demonstrates that Self Forcing achieves superior quality given same wall-clock training budgets compared to both alternative approaches. Each Self Forcing experiment with DMD converges in approximately 1.5 hours on 64 H100 GPUs.

This counter-intuitive result stems from two key factors: First, while Self Forcing performs sequential rollout, it still processes all tokens within each individual frame/chunk in parallel, maintaining high GPU utilization during training. Second, TF and DF require specialized attention masking patterns to enforce causal dependencies, introducing additional computational overhead even with specialized implementations like FlexAttention~\cite{Dong2024FlexAA}.
On the other hand, Self Forcing always uses full attention during training and can leverage highly optimized attention kernels such as FlashAttention-3~\cite{shah2024flashattention}.

\section{Discussion}
\label{sec:discussion}
In this section, we examine the broader implications of our results, discuss additional perspectives, and outline potential directions for future research.

\noindent\textbf{Fundamental limitation of the parallelizable training paradigm.} Parallelizable training has been pivotal to transformers' success by enabling efficient scaling. However, this parallelism introduces fundamental limitations.  Prior research~\cite{merrill2023parallelism} demonstrates that parallel architectures inherently limit expressiveness in sequential state-tracking problems. Our work highlights another critical limitation: parallelizable training paradigms creates misalignment between training and inference distributions, leading to the accumulation of errors over time.
We advocate a new paradigm of \textit{parallel pre-training} and \textit{sequential post-training} that combines the best of both worlds. While this paradigm shift is gaining momentum in language modeling through reinforcement learning~\cite{guo2025deepseek}, our work represents the first step towards this direction for the video domain. We believe our framework is general and can be applied to other sequence domains, especially where the data is continuous.

\noindent\textbf{Interplay between AR, Diffusion, and GANs.}
Autoregressive models, diffusion models, and GANs have traditionally been viewed as distinct paradigms in generative modeling. Our work highlights their complementary nature and demonstrates how they can be effectively integrated. Specifically, autoregressive and diffusion models provide complementary ways to factorize distributions~(chain-rule vs. latent-variable), which can be composed in a nested manner. The core idea behind GANs—matching the distribution of an implicit generator to the target distribution by drawing samples from the implicit generator—can be employed to train a generator powered by autoregressive-diffusion factorization.

\noindent\textbf{Limitation and future directions.}
While our method effectively mitigates error accumulation within the training context length, quality degradation remains observable when generating videos substantially longer than those seen during training. Additionally, our gradient truncation strategies—while necessary for memory efficiency—may limit the model's ability to learn long-range dependencies. Future work could explore both improved extrapolation techniques and inherently recurrent architectures like state-space models~\cite{gumamba,causalmamba} that better balance memory efficiency with long-context modeling.

\section*{Acknowledgments}
We thank Tianwei Yin, Beidi Chen, Kaiwen Zheng, Kai Zhang, Gaurav Parmar, Yi Gu, Sai Bi, and Jianming Zhang for valuable discussions. G. He and M. Zhou acknowledge the support of NSF-IIS 2212418 and NIH-R37
CA271186.

{
  \small
  \bibliographystyle{plain}
  \bibliography{main}

@String(CVPR= {IEEE Conf. Comput. Vis. Pattern Recog.})

@String(ICCV= {Int. Conf. Comput. Vis.})

@String(ECCV= {Eur. Conf. Comput. Vis.})

@String(NIPS= {Adv. Neural Inform. Process. Syst.})

@String(ICLR = {Int. Conf. Learn. Represent.})

@String(AAAI = {AAAI})

@String(CVPR  = {CVPR})

@String(ICCV  = {ICCV})

@String(ECCV  = {ECCV})

@String(NIPS  = {NeurIPS})

@String(NeurIPS  = {NeurIPS})

@String(ICLR  = {ICLR})

@String(ICML = {ICML})

@inproceedings{jenni2019stabilizing,
  title={On stabilizing generative adversarial training with noise},
  author={Jenni, Simon and Favaro, Paolo},
  booktitle=CVPR,
  year={2019}
}

@inproceedings{valevski2025diffusion,
  title={Diffusion models are real-time game engines},
  author={Valevski, Dani and Leviathan, Yaniv and Arar, Moab and Fruchter, Shlomi},
  booktitle=ICLR,
  year={2025}
}

@inproceedings{ho2022video,
  title={Video diffusion models},
  author={Ho, Jonathan and Salimans, Tim and Gritsenko, Alexey and Chan, William and Norouzi, Mohammad and Fleet, David J},
  booktitle=NIPS,
  year={2022}
}

@misc{videoworldsimulators2024,
  title={Video generation models as world simulators},
  author={Tim Brooks and Bill Peebles and Connor Holmes and Will DePue and Yufei Guo and Li Jing and David Schnurr and Joe Taylor and Troy Luhman and Eric Luhman and Clarence Ng and Ricky Wang and Aditya Ramesh},
  year={2024},
  url={https://openai.com/research/video-generation-models-as-world-simulators},
}

@article{polyak2024movie,
  title={Movie gen: A cast of media foundation models},
  author={Polyak, Adam and Zohar, Amit and Brown, Andrew and Tjandra, Andros and Sinha, Animesh and Lee, Ann and Vyas, Apoorv and Shi, Bowen and Ma, Chih-Yao and Chuang, Ching-Yao and others},
  journal={arXiv preprint arXiv:2410.13720},
  year={2024}
}

@inproceedings{yang2025cogvideox,
  title={Cogvideox: Text-to-video diffusion models with an expert transformer},
  author={Yang, Zhuoyi and Teng, Jiayan and Zheng, Wendi and Ding, Ming and Huang, Shiyu and Xu, Jiazheng and Yang, Yuanming and Hong, Wenyi and Zhang, Xiaohan and Feng, Guanyu and others},
  booktitle=ICLR,
  year={2025}
}

@article{kong2024hunyuanvideo,
  title={Hunyuanvideo: A systematic framework for large video generative models},
  author={Kong, Weijie and Tian, Qi and Zhang, Zijian and Min, Rox and Dai, Zuozhuo and Zhou, Jin and Xiong, Jiangfeng and Li, Xin and Wu, Bo and Zhang, Jianwei and others},
  journal={arXiv preprint arXiv:2412.03603},
  year={2024}
}

@inproceedings{liu2021infinite,
  title={Infinite nature: Perpetual view generation of natural scenes from a single image},
  author={Liu, Andrew and Tucker, Richard and Jampani, Varun and Makadia, Ameesh and Snavely, Noah and Kanazawa, Angjoo},
  booktitle=ICCV,
  year={2021}
}

@inproceedings{villegas2023phenaki,
  title={Phenaki: Variable Length Video Generation from Open Domain Textual Descriptions},
  author={Villegas, R and Moraldo, H and Castro, S and Babaeizadeh, M and Zhang, H and Kunze, J and Kindermans, PJ and Saffar, MT and Erhan, D},
  booktitle=ICLR,
  year={2023}
}

@inproceedings{gupta2024photorealistic,
  title={Photorealistic video generation with diffusion models},
  author={Gupta, Agrim and Yu, Lijun and Sohn, Kihyuk and Gu, Xiuye and Hahn, Meera and Li, Fei-Fei and Essa, Irfan and Jiang, Lu and Lezama, Jos{\'e}},
  booktitle=ECCV,
  year={2024}
}

@article{wang2025wan,
  title={Wan: Open and Advanced Large-Scale Video Generative Models},
  author={Wang, Ang and Ai, Baole and Wen, Bin and Mao, Chaojie and Xie, Chen-Wei and Chen, Di and Yu, Feiwu and Zhao, Haiming and Yang, Jianxiao and Zeng, Jianyuan and others},
  journal={arXiv preprint arXiv:2503.20314},
  year={2025}
}

@inproceedings{weissenborn2020scaling,
  title={Scaling Autoregressive Video Models},
  author={Weissenborn, Dirk and T{\"a}ckstr{\"o}m, Oscar and Uszkoreit, Jakob},
  booktitle=ICLR,
  year={2020}
}

@inproceedings{kondratyuk2024videopoet,
  title={VideoPoet: A Large Language Model for Zero-Shot Video Generation},
  author={Kondratyuk, Dan and Yu, Lijun and Gu, Xiuye and Lezama, Jose and Huang, Jonathan and Schindler, Grant and Hornung, Rachel and Birodkar, Vighnesh and Yan, Jimmy and Chiu, Ming-Chang and others},
  booktitle=ICML,
  year={2024}
}

@inproceedings{deng2024nova,
  title={Autoregressive Video Generation without Vector Quantization},
  author={Deng, Haoge and Pan, Ting and Diao, Haiwen and Luo, Zhengxiong and Cui, Yufeng and Lu, Huchuan and Shan, Shiguang and Qi, Yonggang and Wang, Xinlong},
  booktitle=ICLR,
  year={2025}
}

@article{yan2021videogpt,
  title={Videogpt: Video generation using vq-vae and transformers},
  author={Yan, Wilson and Zhang, Yunzhi and Abbeel, Pieter and Srinivas, Aravind},
  journal={arXiv preprint arXiv:2104.10157},
  year={2021}
}

@article{wang2024loong,
  title={Loong: Generating minute-level long videos with autoregressive language models},
  author={Wang, Yuqing and Xiong, Tianwei and Zhou, Daquan and Lin, Zhijie and Zhao, Yang and Kang, Bingyi and Feng, Jiashi and Liu, Xihui},
  journal={arXiv preprint arXiv:2410.02757},
  year={2024}
}

@inproceedings{Bruce2024GenieGI,
  title={Genie: Generative interactive environments},
  author={Bruce, Jake and Dennis, Michael D and Edwards, Ashley and Parker-Holder, Jack and Shi, Yuge and Hughes, Edward and Lai, Matthew and Mavalankar, Aditi and Steigerwald, Richie and Apps, Chris and others},
  booktitle=ICML,
  year={2024}
}

@article{hu2024acdit,
  title={ACDiT: Interpolating Autoregressive Conditional Modeling and Diffusion Transformer},
  author={Hu, Jinyi and Hu, Shengding and Song, Yuxuan and Huang, Yufei and Wang, Mingxuan and Zhou, Hao and Liu, Zhiyuan and Ma, Wei-Ying and Sun, Maosong},
  journal={arXiv preprint arXiv:2412.07720},
  year={2024}
}

@inproceedings{chen2024diffusion,
  title={Diffusion forcing: Next-token prediction meets full-sequence diffusion},
  author={Chen, Boyuan and Mart{\'\i} Mons{\'o}, Diego and Du, Yilun and Simchowitz, Max and Tedrake, Russ and Sitzmann, Vincent},
  booktitle=NIPS,
  year={2024}
}

@inproceedings{yin2025causvid,
  title={From Slow Bidirectional to Fast Autoregressive Video Diffusion Models},
  author={Yin, Tianwei and Zhang, Qiang and Zhang, Richard and Freeman, William T and Durand, Fredo and Shechtman, Eli and Huang, Xun},
  booktitle=CVPR,
  year={2025}
}

@inproceedings{jin2024pyramidal,
  title={Pyramidal Flow Matching for Efficient Video Generative Modeling},
  author={Jin, Yang and Sun, Zhicheng and Li, Ningyuan and Xu, Kun and Jiang, Hao and Zhuang, Nan and Huang, Quzhe and Song, Yang and Mu, Yadong and Lin, Zhouchen},
  booktitle=ICLR,
  year={2025}
}

@article{gu2025long,
  title={Long-Context Autoregressive Video Modeling with Next-Frame Prediction},
  author={Gu, Yuchao and Mao, Weijia and Shou, Mike Zheng},
  journal={arXiv preprint arXiv:2503.19325},
  year={2025}
}

@article{ren2025next,
  title={Next Block Prediction: Video Generation via Semi-Auto-Regressive Modeling},
  author={Ren, Shuhuai and Ma, Shuming and Sun, Xu and Wei, Furu},
  journal={arXiv preprint arXiv:2502.07737},
  year={2025}
}

@misc{parkerholder2024genie2,
  title         = {Genie 2: A Large-Scale Foundation World Model},
  author        = {Jack Parker-Holder and Philip Ball and Jake Bruce and Vibhavari Dasagi and Kristian Holsheimer and Christos Kaplanis and Alexandre Moufarek and Guy Scully and Jeremy Shar and Jimmy Shi and Stephen Spencer and Jessica Yung and Michael Dennis and Sultan Kenjeyev and Shangbang Long and Vlad Mnih and Harris Chan and Maxime Gazeau and Bonnie Li and Fabio Pardo and Luyu Wang and Lei Zhang and Frederic Besse and Tim Harley and Anna Mitenkova and Jane Wang and Jeff Clune and Demis Hassabis and Raia Hadsell and Adrian Bolton and Satinder Singh and Tim Rockt{\"a}schel},
  year          = {2024},
  url           = {https://deepmind.google/discover/blog/genie-2-a-large-scale-foundation-world-model/}
}

@misc{oasis2024,
  author    = {Decart, Julian and Quevedo, Quinn and McIntyre, Spruce and Campbell, Xinlei and Chen, Robert and Wachen},
  title     = {Oasis: A Universe in a Transformer},
  year      = {2024},
  url       = {https://oasis-model.github.io/}
}

@article{yu2025gamefactory,
  title={GameFactory: Creating New Games with Generative Interactive Videos},
  author={Yu, Jiwen and Qin, Yiran and Wang, Xintao and Wan, Pengfei and Zhang, Di and Liu, Xihui},
  journal={arXiv preprint arXiv:2501.08325},
  year={2025}
}

@article{song2025history,
  title={History-Guided Video Diffusion},
  author={Song, Kiwhan and Chen, Boyuan and Simchowitz, Max and Du, Yilun and Tedrake, Russ and Sitzmann, Vincent},
  journal={arXiv preprint arXiv:2502.06764},
  year={2025}
}

@inproceedings{weng2024art,
  title={Art-v: Auto-regressive text-to-video generation with diffusion models},
  author={Weng, Wenming and Feng, Ruoyu and Wang, Yanhui and Dai, Qi and Wang, Chunyu and Yin, Dacheng and Zhao, Zhiyuan and Qiu, Kai and Bao, Jianmin and Yuan, Yuhui and others},
  booktitle=CVPR,
  year={2024}
}

@inproceedings{van2017neural,
  title={Neural discrete representation learning},
  author={Van Den Oord, Aaron and Vinyals, Oriol and others},
  booktitle=NIPS,
  year={2017}
}

@inproceedings{peebles2022scalable,
  title={Scalable diffusion models with transformers},
  author={Peebles, William S and Xie, Saining},
  booktitle=ICCV,
  year={2023}
}

@inproceedings{hong2022cogvideo,
  title={Cogvideo: Large-scale pretraining for text-to-video generation via transformers},
  author={Hong, Wenyi and Ding, Ming and Zheng, Wendi and Liu, Xinghan and Tang, Jie},
  booktitle=ICLR,
  year={2023}
}

@article{blattmann2023stable,
  title={Stable video diffusion: Scaling latent video diffusion models to large datasets},
  author={Blattmann, Andreas and Dockhorn, Tim and Kulal, Sumith and Mendelevitch, Daniel and Kilian, Maciej and Lorenz, Dominik and Levi, Yam and English, Zion and Voleti, Vikram and Letts, Adam and others},
  journal={arXiv preprint arXiv:2311.15127},
  year={2023}
}

@article{hacohen2024ltx,
  title={Ltx-video: Realtime video latent diffusion},
  author={HaCohen, Yoav and Chiprut, Nisan and Brazowski, Benny and Shalem, Daniel and Moshe, Dudu and Richardson, Eitan and Levin, Eran and Shiran, Guy and Zabari, Nir and Gordon, Ori and others},
  journal={arXiv preprint arXiv:2501.00103},
  year={2024}
}

@inproceedings{zhao2025real,
  title={Real-time video generation with pyramid attention broadcast},
  author={Zhao, Xuanlei and Jin, Xiaolong and Wang, Kai and You, Yang},
  booktitle=ICLR,
  year={2025}
}

@article{Ho2022ImagenVH,
  title={Imagen Video: High Definition Video Generation with Diffusion Models},
  author={Jonathan Ho and William Chan and Chitwan Saharia and Jay Whang and Ruiqi Gao and Alexey A. Gritsenko and Diederik P. Kingma and Ben Poole and Mohammad Norouzi and David J. Fleet and Tim Salimans},
  journal={ArXiv},
  year={2022},
  volume={abs/2210.02303},
  url={https://api.semanticscholar.org/CorpusID:252715883}
}

@inproceedings{ge2022long,
  title={Long video generation with time-agnostic vqgan and time-sensitive transformer},
  author={Ge, Songwei and Hayes, Thomas and Yang, Harry and Yin, Xi and Pang, Guan and Jacobs, David and Huang, Jia-Bin and Parikh, Devi},
  booktitle=ECCV,
  year={2022}
}

@inproceedings{blattmann2023align,
  title={Align your latents: High-resolution video synthesis with latent diffusion models},
  author={Blattmann, Andreas and Rombach, Robin and Ling, Huan and Dockhorn, Tim and Kim, Seung Wook and Fidler, Sanja and Kreis, Karsten},
  booktitle=CVPR,
  year={2023}
}

@article{gao2024ca2,
  title={Ca2-VDM: Efficient Autoregressive Video Diffusion Model with Causal Generation and Cache Sharing},
  author={Gao, Kaifeng and Shi, Jiaxin and Zhang, Hanwang and Wang, Chunping and Xiao, Jun and Chen, Long},
  journal={arXiv preprint arXiv:2411.16375},
  year={2024}
}

@inproceedings{li2025arlon,
  title={Arlon: Boosting diffusion transformers with autoregressive models for long video generation},
  author={Li, Zongyi and Hu, Shujie and Liu, Shujie and Zhou, Long and Choi, Jeongsoo and Meng, Lingwei and Guo, Xun and Li, Jinyu and Ling, Hefei and Wei, Furu},
  booktitle=ICLR,
  year={2025}
}

@inproceedings{sun2025ar,
  title={AR-Diffusion: Asynchronous Video Generation with Auto-Regressive Diffusion},
  author={Sun, Mingzhen and Wang, Weining and Li, Gen and Liu, Jiawei and Sun, Jiahui and Feng, Wanquan and Lao, Shanshan and Zhou, SiYu and He, Qian and Liu, Jing},
  booktitle=CVPR,
  year={2025}
}

@article{chen2025skyreels,
  title={SkyReels-V2: Infinite-length Film Generative Model},
  author={Chen, Guibin and Lin, Dixuan and Yang, Jiangping and Lin, Chunze and Zhu, Juncheng and Fan, Mingyuan and Zhang, Hao and Chen, Sheng and Chen, Zheng and Ma, Chengchen and others},
  journal={arXiv preprint arXiv:2504.13074},
  year={2025}
}

@article{deng2024causal,
  title={Causal diffusion transformers for generative modeling},
  author={Deng, Chaorui and Zhu, Deyao and Li, Kunchang and Guang, Shi and Fan, Haoqi},
  journal={arXiv preprint arXiv:2412.12095},
  year={2024}
}

@article{liu2024mardini,
  title={Mardini: Masked autoregressive diffusion for video generation at scale},
  author={Liu, Haozhe and Liu, Shikun and Zhou, Zijian and Xu, Mengmeng and Xie, Yanping and Han, Xiao and P{\'e}rez, Juan C and Liu, Ding and Kahatapitiya, Kumara and Jia, Menglin and others},
  journal={arXiv preprint arXiv:2410.20280},
  year={2024}
}

@inproceedings{streamingllm,
  title={Efficient Streaming Language Models with Attention Sinks},
  author={Xiao, Guangxuan and Tian, Yuandong and Chen, Beidi and Han, Song and Lewis, Mike},
  booktitle=ICLR,
  year={2024}
}

@article{schmidt2019generalization,
  title={Generalization in Generation: A closer look at Exposure Bias},
  author={Schmidt, Florian},
  journal={EMNLP-IJCNLP 2019},
  pages={157},
  year={2019}
}

@inproceedings{ning2024elucidating,
  title={Elucidating the Exposure Bias in Diffusion Models},
  author={Ning, Mang and Li, Mingxiao and Su, Jianlin and Salah, Albert Ali and Ertugrul, Itir Onal},
  booktitle=ICLR,
  year={2024}
}

@article{guo2025long,
  title={Long context tuning for video generation},
  author={Guo, Yuwei and Yang, Ceyuan and Yang, Ziyan and Ma, Zhibei and Lin, Zhijie and Yang, Zhenheng and Lin, Dahua and Jiang, Lu},
  journal={arXiv preprint arXiv:2503.10589},
  year={2025}
}

@inproceedings{lamb2016professor,
  title={Professor forcing: A new algorithm for training recurrent networks},
  author={Lamb, Alex M and ALIAS PARTH GOYAL, Anirudh Goyal and Zhang, Ying and Zhang, Saizheng and Courville, Aaron C and Bengio, Yoshua},
  booktitle=NIPS,
  year={2016}
}

@inproceedings{ranzato2016sequence,
  title={Sequence level training with recurrent neural networks},
  author={Ranzato, Marc’Aurelio and Chopra, Sumit and Auli, Michael and Zaremba, Wojciech},
  booktitle=ICLR,
  year={2016}
}

@inproceedings{yu2017seqgan,
  title={Seqgan: Sequence generative adversarial nets with policy gradient},
  author={Yu, Lantao and Zhang, Weinan and Wang, Jun and Yu, Yong},
  booktitle=AAAI,
  year={2017}
}

@inproceedings{liang2025looking,
  title={Looking backward: Streaming video-to-video translation with feature banks},
  author={Liang, Feng and Kodaira, Akio and Xu, Chenfeng and Tomizuka, Masayoshi and Keutzer, Kurt and Marculescu, Diana},
  booktitle=ICLR,
  year={2025}
}

@article{chen2024streaming,
  title={Streaming video diffusion: Online video editing with diffusion models},
  author={Chen, Feng and Yang, Zhen and Zhuang, Bohan and Wu, Qi},
  journal={arXiv preprint arXiv:2405.19726},
  year={2024}
}

@inproceedings{yang2023learning,
  title={Learning interactive real-world simulators},
  author={Yang, Mengjiao and Du, Yilun and Ghasemipour, Kamyar and Tompson, Jonathan and Schuurmans, Dale and Abbeel, Pieter},
  booktitle=ICLR,
  year={2024}
}

@inproceedings{yu2024learning,
  title={Learning Visual Parkour from Generated Images},
  author={Yu, Alan and Yang, Ge and Choi, Ran and Ravan, Yajvan and Leonard, John and Isola, Phillip},
  booktitle={CoRL},
  year={2024}
}

@article{li2025unified,
  title={Unified Video Action Model},
  author={Li, Shuang and Gao, Yihuai and Sadigh, Dorsa and Song, Shuran},
  journal={arXiv preprint arXiv:2503.00200},
  year={2025}
}

@inproceedings{goodfellow2014generative,
  title={Generative adversarial nets},
  author={Goodfellow, Ian J and Pouget-Abadie, Jean and Mirza, Mehdi and Xu, Bing and Warde-Farley, David and Ozair, Sherjil and Courville, Aaron and Bengio, Yoshua},
  booktitle=NeurIPS,
  year={2014}
}

@inproceedings{wu2023ar,
  title={Ar-diffusion: Auto-regressive diffusion model for text generation},
  author={Wu, Tong and Fan, Zhihao and Liu, Xiao and Zheng, Hai-Tao and Gong, Yeyun and Jiao, Jian and Li, Juntao and Guo, Jian and Duan, Nan and Chen, Weizhu and others},
  booktitle=NeurIPS,
  year={2023}
}

@article{zhang2025generative,
  title={Generative Pre-trained Autoregressive Diffusion Transformer},
  author={Zhang, Yuan and Jiang, Jiacheng and Ma, Guoqing and Lu, Zhiying and Huang, Haoyang and Yuan, Jianlong and Duan, Nan},
  journal={arXiv preprint arXiv:2505.07344},
  year={2025}
}

@inproceedings{zhou2025taming,
  title={Taming Teacher Forcing for Masked Autoregressive Video Generation},
  author={Zhou, Deyu and Sun, Quan and Peng, Yuang and Yan, Kun and Dong, Runpei and Wang, Duomin and Ge, Zheng and Duan, Nan and Zhang, Xiangyu and Ni, Lionel M and others},
  booktitle=CVPR,
  year={2025}
}

@article{liu2024redefining,
  title={Redefining Temporal Modeling in Video Diffusion: The Vectorized Timestep Approach},
  author={Liu, Yaofang and Ren, Yumeng and Cun, Xiaodong and Artola, Aitor and Liu, Yang and Zeng, Tieyong and Chan, Raymond H and Morel, Jean-michel},
  journal={arXiv preprint arXiv:2410.03160},
  year={2024}
}

@article{liu2024autoregressive,
  title={Autoregressive diffusion transformer for text-to-speech synthesis},
  author={Liu, Zhijun and Wang, Shuai and Inoue, Sho and Bai, Qibing and Li, Haizhou},
  journal={arXiv preprint arXiv:2406.05551},
  year={2024}
}

@inproceedings{arriola2025block,
  title={Block diffusion: Interpolating between autoregressive and diffusion language models},
  author={Arriola, Marianne and Gokaslan, Aaron and Chiu, Justin T and Yang, Zhihan and Qi, Zhixuan and Han, Jiaqi and Sahoo, Subham Sekhar and Kuleshov, Volodymyr},
  booktitle=ICLR,
  year={2025}
}

@inproceedings{li2024autoregressive,
  title={Autoregressive image generation without vector quantization},
  author={Li, Tianhong and Tian, Yonglong and Li, He and Deng, Mingyang and He, Kaiming},
  booktitle=NeurIPS,
  year={2024}
}

@inproceedings{yu2023language,
  title={Language Model Beats Diffusion--Tokenizer is Key to Visual Generation},
  author={Yu, Lijun and Lezama, Jos{\'e} and Gundavarapu, Nitesh B and Versari, Luca and Sohn, Kihyuk and Minnen, David and Cheng, Yong and Birodkar, Vighnesh and Gupta, Agrim and Gu, Xiuye and others},
  booktitle=ICLR,
  year={2024}
}

@inproceedings{kingma2013auto,
  title={Auto-Encoding Variational Bayes},
  author={Kingma, Diederik P and Welling, Max},
  booktitle=ICLR,
  year={2014}
}

@article{Dong2024FlexAA,
  title={Flex Attention: A Programming Model for Generating Optimized Attention Kernels},
  author={Juechu Dong and Boyuan Feng and Driss Guessous and Yanbo Liang and Horace He},
  journal={ArXiv},
  year={2024},
  volume={abs/2412.05496},
  url={https://api.semanticscholar.org/CorpusID:274598006}
}

@inproceedings{song2023consistency,
  title={Consistency Models},
  author={Song, Yang and Dhariwal, Prafulla and Chen, Mark and Sutskever, Ilya},
  booktitle=ICML,
  year={2023}
}

@article{wang2023prolificdreamer,
  title={Prolificdreamer: High-fidelity and diverse text-to-3d generation with variational score distillation},
  author={Wang, Zhengyi and Lu, Cheng and Wang, Yikai and Bao, Fan and Li, Chongxuan and Su, Hang and Zhu, Jun},
  journal=NeurIPS,
  year={2023}
}

@inproceedings{
luo2023diffinstruct,
title={Diff-Instruct: A Universal Approach for Transferring Knowledge From Pre-trained Diffusion Models},
author={Weijian Luo and Tianyang Hu and Shifeng Zhang and Jiacheng Sun and Zhenguo Li and Zhihua Zhang},
booktitle=NeurIPS,
year={2023},
}

@inproceedings{yin2024one,
  title={One-step diffusion with distribution matching distillation},
  author={Yin, Tianwei and Gharbi, Micha{\"e}l and Zhang, Richard and Shechtman, Eli and Durand, Fredo and Freeman, William T and Park, Taesung},
  booktitle=CVPR,
  year={2024}
}

@article{yin2024improved,
  title={Improved distribution matching distillation for fast image synthesis},
  author={Yin, Tianwei and Gharbi, Micha{\"e}l and Park, Taesung and Zhang, Richard and Shechtman, Eli and Durand, Fredo and Freeman, Bill},
  journal=NeurIPS,
  year={2024}
}

@inproceedings{zhou2024score,
  title={Score identity distillation: Exponentially fast distillation of pretrained diffusion models for one-step generation},
  author={Zhou, Mingyuan and Zheng, Huangjie and Wang, Zhendong and Yin, Mingzhang and Huang, Hai},
  booktitle=ICML,
  year={2024}
}

@article{luo2024one,
  title={One-step diffusion distillation through score implicit matching},
  author={Luo, Weijian and Huang, Zemin and Geng, Zhengyang and Kolter, J Zico and Qi, Guo-jun},
  journal=NeurIPS,
  year={2024}
}

@article{huang2024flow,
  title={Flow generator matching},
  author={Huang, Zemin and Geng, Zhengyang and Luo, Weijian and Qi, Guo-jun},
  journal={arXiv preprint arXiv:2410.19310},
  year={2024}
}

@inproceedings{zhou2025adversarial,
  title={Adversarial Score identity Distillation: Rapidly Surpassing the Teacher in One Step},
  author={Zhou, Mingyuan and Zheng, Huangjie and Gu, Yi and Wang, Zhendong and Huang, Hai},
  booktitle=ICLR,
  year={2025}
}

@inproceedings{sauer2024fast,
  title={Fast high-resolution image synthesis with latent adversarial diffusion distillation},
  author={Sauer, Axel and Boesel, Frederic and Dockhorn, Tim and Blattmann, Andreas and Esser, Patrick and Rombach, Robin},
  booktitle={SIGGRAPH Asia 2024 Conference Papers},
  pages={1--11},
  year={2024}
}

@inproceedings{villegas2017decomposing,
  title={Decomposing motion and content for natural video sequence prediction},
  author={Villegas, Ruben and Yang, Jimei and Hong, Seunghoon and Lin, Xunyu and Lee, Honglak},
  booktitle=ICLR,
  year={2017}
}

@inproceedings{denton2017unsupervised,
  title={Unsupervised learning of disentangled representations from video},
  author={Denton, Emily L and others},
  booktitle=NIPS,
  year={2017}
}

@inproceedings{tulyakov2018mocogan,
  title={Mocogan: Decomposing motion and content for video generation},
  author={Tulyakov, Sergey and Liu, Ming-Yu and Yang, Xiaodong and Kautz, Jan},
  booktitle=CVPR,
  year={2018}
}

@inproceedings{li2022infinitenature,
  title={Infinitenature-zero: Learning perpetual view generation of natural scenes from single images},
  author={Li, Zhengqi and Wang, Qianqian and Snavely, Noah and Kanazawa, Angjoo},
  booktitle=ECCV,
  year={2022}
}

@inproceedings{saito2017temporal,
  title={Temporal generative adversarial nets with singular value clipping},
  author={Saito, Masaki and Matsumoto, Eiichi and Saito, Shunta},
  booktitle=ICCV,
  year={2017}
}

@article{vondrick2016generating,
  title={Generating videos with scene dynamics},
  author={Vondrick, Carl and Pirsiavash, Hamed and Torralba, Antonio},
  journal=NIPS,
  year={2016}
}

@article{brooks2022generating,
  title={Generating long videos of dynamic scenes},
  author={Brooks, Tim and Hellsten, Janne and Aittala, Miika and Wang, Ting-Chun and Aila, Timo and Lehtinen, Jaakko and Liu, Ming-Yu and Efros, Alexei and Karras, Tero},
  journal=NIPS,
  year={2022}
}

@inproceedings{ruhe2024rolling,
  title={Rolling diffusion models},
  author={Ruhe, David and Heek, Jonathan and Salimans, Tim and Hoogeboom, Emiel},
  booktitle=ICML,
  year={2024}
}

@inproceedings{kim2025fifo,
  title={FIFO-Diffusion: Generating Infinite Videos from Text without Training},
  author={Kim, Jihwan and Kang, Junoh and Choi, Jinyoung and Han, Bohyung},
  booktitle=NIPS,
  year={2024}
}

@article{xie2024progressive,
  title={Progressive autoregressive video diffusion models},
  author={Xie, Desai and Xu, Zhan and Hong, Yicong and Tan, Hao and Liu, Difan and Liu, Feng and Kaufman, Arie and Zhou, Yang},
  journal={arXiv preprint arXiv:2410.08151},
  year={2024}
}

@article{zhang2025packing,
  title={Packing Input Frame Context in Next-Frame Prediction Models for Video Generation},
  author={Zhang, Lvmin and Agrawala, Maneesh},
  journal={arXiv preprint arXiv:2504.12626},
  year={2025}
}

@misc{magi1,
      title={MAGI-1: Autoregressive Video Generation at Scale},
      author={Sand-AI},
      year={2025},
      url={https://static.magi.world/static/files/MAGI_1.pdf},
}

@inproceedings{zhou2025transfusion,
  title={Transfusion: Predict the Next Token and Diffuse Images with One Multi-Modal Model},
  author={Zhou, Chunting and Yu, Lili and Babu, Arun and Tirumala, Kushal and Yasunaga, Michihiro and Shamis, Leonid and Kahn, Jacob and Ma, Xuezhe and Zettlemoyer, Luke and Levy, Omer},
  booktitle=ICLR,
  year={2025}
}

@article{mo2025xfusionintroducingnewmodality,
  title={X-Fusion: Introducing New Modality to Frozen Large Language Models},
  author={Mo, Sicheng and Nguyen, Thao and Huang, Xun and Iyer, Siddharth Srinivasan and Li, Yijun and Liu, Yuchen and Tandon, Abhishek and Shechtman, Eli and Singh, Krishna Kumar and Lee, Yong Jae and others},
  journal={arXiv preprint arXiv:2504.20996},
  year={2025}
}

@inproceedings{kingma2021variational,
  title={Variational diffusion models},
  author={Kingma, Diederik and Salimans, Tim and Poole, Ben and Ho, Jonathan},
  booktitle=NeurIPS,
  year={2021}
}

@inproceedings{song2021maximum,
  title={Maximum likelihood training of score-based diffusion models},
  author={Song, Yang and Durkan, Conor and Murray, Iain and Ermon, Stefano},
  booktitle=NeurIPS,
  year={2021}
}

@article{lin2025diffusion,
  title={Diffusion adversarial post-training for one-step video generation},
  author={Lin, Shanchuan and Xia, Xin and Ren, Yuxi and Yang, Ceyuan and Xiao, Xuefeng and Jiang, Lu},
  journal={arXiv preprint arXiv:2501.08316},
  year={2025}
}

@inproceedings{mao2025osv,
  title={Osv: One step is enough for high-quality image to video generation},
  author={Mao, Xiaofeng and Jiang, Zhengkai and Wang, Fu-Yun and Zhang, Jiangning and Chen, Hao and Chi, Mingmin and Wang, Yabiao and Luo, Wenhan},
  booktitle=CVPR,
  year={2025}
}

@inproceedings{wu2025snapgen,
  title={SnapGen-V: Generating a Five-Second Video within Five Seconds on a Mobile Device},
  author={Wu, Yushu and Zhang, Zhixing and Li, Yanyu and Xu, Yanwu and Kag, Anil and Sui, Yang and Coskun, Huseyin and Ma, Ke and Lebedev, Aleksei and Hu, Ju and others},
  booktitle=CVPR,
  year={2025}
}

@inproceedings{zhang2024sf,
  title={Sf-v: Single forward video generation model},
  author={Zhang, Zhixing and Li, Yanyu and Wu, Yushu and Kag, Anil and Skorokhodov, Ivan and Menapace, Willi and Siarohin, Aliaksandr and Cao, Junli and Metaxas, Dimitris and Tulyakov, Sergey and others},
  booktitle=NIPS,
  year={2024}
}

@inproceedings{wangvidprom,
  title={VidProM: A Million-scale Real Prompt-Gallery Dataset for Text-to-Video Diffusion Models},
  author={Wang, Wenhao and Yang, Yi},
  booktitle=NIPS,
  year={2024},
}

@inproceedings{huang2023vbench,
      title={{VBench}: Comprehensive Benchmark Suite for Video Generative Models},
      author={Huang, Ziqi and He, Yinan and Yu, Jiashuo and Zhang, Fan and Si, Chenyang and Jiang, Yuming and Zhang, Yuanhan and Wu, Tianxing and Jin, Qingyang and Chanpaisit, Nattapol and Wang, Yaohui and Chen, Xinyuan and Wang, Limin and Lin, Dahua and Qiao, Yu and Liu, Ziwei},
      booktitle=CVPR,
      year={2024}
}

@inproceedings{lipmanflow,
  title={Flow Matching for Generative Modeling},
  author={Lipman, Yaron and Chen, Ricky TQ and Ben-Hamu, Heli and Nickel, Maximilian and Le, Matthew},
  booktitle=ICLR,
  year={2023}
}

@inproceedings{liu2022flow,
  title={Flow straight and fast: Learning to generate and transfer data with rectified flow},
  author={Liu, Xingchao and Gong, Chengyue and Liu, Qiang},
  booktitle=ICLR,
  year={2023}
}

@inproceedings{huang2024gan,
  title={The GAN is dead; long live the GAN! A Modern GAN Baseline},
  author={Huang, Nick and Gokaslan, Aaron and Kuleshov, Volodymyr and Tompkin, James},
  booktitle=NIPS,
  year={2024}
}

@inproceedings{mescheder2018training,
  title={Which training methods for GANs do actually converge?},
  author={Mescheder, Lars and Geiger, Andreas and Nowozin, Sebastian},
  booktitle=ICML,
  year={2018}
}

@inproceedings{jolicoeurrelativistic,
  title={The relativistic discriminator: a key element missing from standard GAN},
  author={Jolicoeur-Martineau, Alexia},
  booktitle=ICLR,
  year={2019}
}

@article{merrill2023parallelism,
  title={The parallelism tradeoff: Limitations of log-precision transformers},
  author={Merrill, William and Sabharwal, Ashish},
  journal={TACL},
  year={2023}
}

@article{guo2025deepseek,
  title={Deepseek-r1: Incentivizing reasoning capability in llms via reinforcement learning},
  author={Guo, Daya and Yang, Dejian and Zhang, Haowei and Song, Junxiao and Zhang, Ruoyu and Xu, Runxin and Zhu, Qihao and Ma, Shirong and Wang, Peiyi and Bi, Xiao and others},
  journal={arXiv preprint arXiv:2501.12948},
  year={2025}
}

@inproceedings{gumamba,
  title={Mamba: Linear-Time Sequence Modeling with Selective State Spaces},
  author={Gu, Albert and Dao, Tri},
  booktitle={COLM},
  year={2024},
}

@article{li4654242latency,
  title={Latency reducing in real-time internet video transport: A survey},
  author={Li, Qing and Tang, Xun and Peng, Junkun and Tan, Yuanzheng and Jiang, Yong},
  journal={SSRN 4654242},
  year={2023}
}

@inproceedings{shah2024flashattention,
  title={Flashattention-3: Fast and accurate attention with asynchrony and low-precision},
  author={Shah, Jay and Bikshandi, Ganesh and Zhang, Ying and Thakkar, Vijay and Ramani, Pradeep and Dao, Tri},
  booktitle=NIPS,
  year={2024}
}

@article{bentaleb2025toward,
  title={Toward one-second latency: Evolution of live media streaming},
  author={Bentaleb, Abdelhak and Lim, May and Akcay, Mehmet N and Begen, Ali C and Hammoudi, Sarra and Zimmermann, Roger},
  journal={IEEE Communications Surveys \& Tutorials},
  year={2025},
  publisher={IEEE}
}

@article{yang2024qwen2,
  title={Qwen2. 5 technical report},
  author={Yang, An and Yang, Baosong and Zhang, Beichen and Hui, Binyuan and Zheng, Bo and Yu, Bowen and Li, Chengyuan and Liu, Dayiheng and Huang, Fei and Wei, Haoran and others},
  journal={arXiv preprint arXiv:2412.15115},
  year={2024}
}

@article{causalmamba,
  title={Long-Context State-Space Video World Models},
  author={Ryan Po and Yotam Nitzan and Richard Zhang and Berlin Chen and Tri Dao and Eli Shechtman and Gordon Wetzstein and Xun Huang},
  journal={arXiv preprint arXiv:2505.20171},
  year={2025}
}

@article{wang2025error,
  title={Error Analyses of Auto-Regressive Video Diffusion Models: A Unified Framework},
  author={Wang, Jing and Zhang, Fengzhuo and Li, Xiaoli and Tan, Vincent YF and Pang, Tianyu and Du, Chao and Sun, Aixin and Yang, Zhuoran},
  journal={arXiv preprint arXiv:2503.10704},
  year={2025}
}

@article{zhang2025test,
  title={Test-Time Training Done Right},
  author={Zhang, Tianyuan and Bi, Sai and Hong, Yicong and Zhang, Kai and Luan, Fujun and Yang, Songlin and Sunkavalli, Kalyan and Freeman, William T and Tan, Hao},
  journal={arXiv preprint arXiv:2505.23884},
  year={2025}
}
}

\appendix

\section{Implementation Details}
\label{app:implementation}

Our implementation is largely based on the open-source code of Wan2.1~\cite{wang2025wan} and CausVid~\cite{yin2025causvid}. The attention implementation of Diffusion Forcing and Teacher Forcing baselines is based on FlexAttention~\cite{Dong2024FlexAA}, while the attention in Self Forcing is based on FlashAttention-3~\cite{shah2024flashattention}.

\paragraph{Noise schedule and model parameterization.}
Following the Wan2.1 series, we adopt the flow matching framework~\cite{lipmanflow, liu2022flow}, with time step shifting $t^\prime(k, t) = (kt / 1000) / (1 + (k-1)(t / 1000)) \cdot 1000$ and a shift factor $k=5$. The forward process is specified as $x_t = \frac{t^\prime}{1000} x + \frac{1 - t^\prime}{1000} \epsilon, \epsilon \sim \mathcal{N}(0, I)$ with $t \in [0, 1000]$.

The data prediction model is given by:
\begin{align}
    G_\theta(x, t, c) = c_{\text{skip}} \cdot \epsilon - c_{\text{out}} \cdot v_\theta(c_{\text{in}} \cdot x_t, c_{\text{noise}}(t^\prime), c).
\end{align}
We keep the preconditioning coefficients the same as the base models' configuration, i.e., $c_{\text{skip}} = c_{\text{in}} = c_{\text{out}} = 1$ and $c_{\text{noise}}(t) = t$.
Our few-step diffusion process employs a uniform 4-step schedule $[t_4, t_3, t_2, t_1]=[1000, 750, 500, 250]$.

\paragraph{Prompt preprocessing.} We use the VidProS subset from VidProM~\cite{wangvidprom}, which contains around 1M semantically unique user-written text-to-video prompts. We filter out prompts that are too short (less than 20 characters), contain command line arguments (e.g., --ar 16:9), or have a NSFW probability greater than 0.01 for any annotated category~(toxicity, obscenity, identity attack, insult, threat, and sexual explicitness). This results in a total of around 250k prompts. We then expand those prompts with \texttt{Qwen/Qwen2.5-7B-Instruct}~\cite{yang2024qwen2}, using the system prompt~(English version) provided in the open-source implementation of Wan2.1~\cite{wang2025wan}. For VBench evaluation, we similarly rewrite the test prompts using \texttt{Qwen/Qwen2.5-7B-Instruct}.
We note that the VBench results of the Wan2.1 base model are also obtained with prompt rewriting, and we report baseline results with prompt rewriting, provided that the model supports such enhancements.

\paragraph{Training details.}
Most of our training runs use 64 NVIDIA GPUs (80GB memory each) with a per-GPU batch size of 1. We implement gradient accumulation for configurations requiring a larger effective batch size than 64. Our DMD training runs take only approximately 1.5 hours to converge, while SiD/GAN training takes 2-3 hours on 64 H100 GPUs.
We initialize the real score network and critic network using the pretrained weights of the base model.
We list all other training configurations, as well as the choice of real score network and critic network for different distribution matching objectives, in Table~\ref{tab:training-hyperparameters}. We describe detailed training configurations for each distribution matching objective below.

For DMD, the gradient of the reverse Kullback-Leibler divergence is given by~\cite{yin2024one, wang2023prolificdreamer, luo2023diffinstruct}:
\begin{align}
\label{eqn:DMD-gradient}
    \nabla_\theta  \mathbb E_{t}[D_\text{KL}(p_{\theta, t} \| p_{\text{data}, t})] = -\mathbb E_{t, \hat x_t \sim q_{t|0}(\hat x_t | \hat x), \hat x \sim p_\theta(\hat x)} \left[ (s_{\text{real}}(\hat x_t, t) - s_{\text{fake}}(\hat x_t, t))\frac{\partial \hat x}{\partial \theta}\right],
\end{align}
where $s_\text{real}(\cdot, t)$ is the score function for $p_{\text{data}, t}$, approximated by a pretrained diffusion model $f_\phi(\cdot, t)$, also referred to as the real score network, and $s_{\text{fake}}(\cdot, t)$ is the score function for $p_{\theta, t}$ and is learned through a critic network $f_\psi(\cdot, t)$ via the standard diffusion loss. The gradient in Eqn.~\eqref{eqn:DMD-gradient} is equivalent to the following loss function:
\begin{align}
    \mathcal L_{\text{DMD}}(\theta) = \mathbb E_{t, \hat x_t, \hat x}\left[ \frac{1}{2}\left\| \hat x - \mathrm{sg}\left[\hat x  - (f_\psi(\hat x_t, t) - f_\phi(\hat x_t, t))\right]\right\|^2\right],
\end{align}
where $\mathrm{sg}[\cdot]$ denotes the stop gradient operator.

Similar to the pipeline of DMD, the SiD loss is given by~\cite{zhou2024score}:
\begin{align}
    \mathcal L_{\text{SiD}}(\theta) = \mathbb E_{t, \hat x_t, \hat x}\left[ (f_\phi(\hat x_t, t) - f_\psi(\hat x_t, t))^T (f_\psi(\hat x_t, t) - \hat x) + (1 - \alpha) \| f_{\phi}(\hat x_t , t) - f_\psi(\hat x_t, t)\|^2 \right],
\end{align}
which can be shown that the case of $\alpha = 0.5$ corresponds the gradient of the Fisher divergence $\mathbb E_{t, p_{\theta, t}}[\| \nabla \log p_{\theta, t}\ - \nabla \log p_{\text{data}, t}\|^2]$ ~\cite{luo2024one, huang2024flow}. Empirically, it is observed that the second term often leads to unstable training and thus $\alpha=1$ is typically adopted for better performance~\cite{zhou2024score, zhou2025adversarial}, which is also followed in this work.


\begin{table}[t]
\caption{Specification of training hyperparameters}
\label{tab:training-hyperparameters}
\vspace{0.5em}
\centering
\renewcommand{\arraystretch}{1.5} 
\resizebox{1\textwidth}{!}{
\begin{tabular}{cccc}
\toprule
Hyperparameters                         & DMD                                                                                  & SiD                                                    & GAN                                                                                 \\
\midrule
Real score network                      & Wan2.1-T2V-14B                                                                       & Wan2.1-T2V-1.3B                                        & N/A                                                                      \\
Real score CFG weight                      & 3.0                                                                       & 3.0                                        & N/A                                                                      \\
Critic network initialization           & Wan2.1-T2V-1.3B                                                                      & Wan2.1-T2V-1.3B                                        & Wan2.1-T2V-1.3B                                                                     \\
Batch size                              & 64                                                                                   & 64                                                     & 768                                                                                  \\
\multirow{2}{*}{Optimizer ($G_\theta$)} & AdamW, $\beta_1=0$, $\beta_2=0.999$,                                                                                & Adam, $\beta_1=0$, $\beta_2=0.999$,                                                    & AdamW, $\beta_1=0$, $\beta_2=0.999$,                                                                                \\
                                        & $\epsilon=1\text{e-8}$,  weight\_decay$=0.01$ & $\epsilon=1\text{e-8}$,  weight\_decay$=0$ & $\epsilon=1\text{e-8}$, weight\_decay$=0.01$ \\
\multirow{2}{*}{Optimizer ($f_\psi$)} & AdamW, $\beta_1=0$, $\beta_2=0.999$,                                                                                & Adam, $\beta_1=0$, $\beta_2=0.999$,                                                    & AdamW, $\beta_1=0$, $\beta_2=0.999$,                                                                                \\
                                        & $\epsilon=1\text{e-8}$,  weight\_decay$=0.01$ & $\epsilon=1\text{e-8}$,  weight\_decay$=0$ & $\epsilon=1\text{e-8}$, weight\_decay$=0.01$ \\
Learning rate ($G_\theta$)              & $2\text{e-6}$                                                                        & $2\text{e-6}$                                          & $2\text{e-6}$                                                                       \\
Learning rate ($f_\psi$)                & $4\text{e-7}$                                                                        & $2\text{e-6}$                                          & $2\text{e-6}$                                                                      \\
Generator/critic update ratio                & $5$                                                                        & $5$                                          & $1$                                                                      \\
EMA decay                & $0.99$                                                                        & $0.99$                                          & $0.99$                                                                      \\
\bottomrule
\end{tabular}
}
\vspace{-1em}
\end{table}

For GAN training, we add additional cross-attention layers and classification heads to the initialized critic network. We employ relativistic loss~\cite{jolicoeurrelativistic} and approximate the regularization terms~(R1 and R2) using finite difference following Seaweed-APT~\cite{lin2025diffusion}. Specifically, we perturb the noisy real/fake data with additional small Gaussian noise and encourage the discriminator output to be similar to the original one. The final training objective is defined as:
\begin{align}
    & \mathcal{L}_{\text{reg}} = \frac{1}{2} \mathbb{E}_{t, x_t, \hat{x}_t, \epsilon, \hat{\epsilon}} \left[ \| f_\psi(x_t) - f_\psi(x_t+\sigma \cdot \epsilon ) \|_2^2 + \| f_\psi(\hat{x}_t) - f_\psi(\hat{x}_t + \sigma \cdot \hat{\epsilon} ) \|_2^2 \right] \\
    & \mathcal{L}_{D}(\psi) = -\mathbb{E}_{t, x_t, \hat{x}_t} \left[ \log \left( \text{sigmoid} \left(f_\psi(x_t) - f_\psi(\hat{x}_t) \right)\right) \right] + \lambda\mathcal{L}_{\text{reg}} \\
    & \mathcal{L}_{G}(\theta) = -\mathbb{E}_{t, x_t, \hat{x}_t} \left[ \log \left( \text{sigmoid} \left(f_\psi(\hat{x}_t) - f_\psi(x_t) \right)\right) \right]
\end{align}
where $x_t\sim p_{\text{data}, t}$, $\hat{x}_t\sim p_{\theta, t}$ are the noisy real and fake data, respectively, $\epsilon$ and $\hat{\epsilon}$ are Gaussian noise sampled from $\mathcal{N}(0, 1)$, and $f_\psi$ is the critic network (discriminator) of GAN. We use $\lambda=30$, $\sigma=0.05$ for all experiments.  For a video generated from the output of the $s$-th step~(see Algorithm~\ref{alg:self_forcing} for details), we find that only sampling $t$ from $[t_{s-1}, t_{s}]$ helps stabilize the training. We also adopt a large batch size of 768 for training stability.

\section{Importance of local attention training in rolling KV cache}
\label{app:qualitative}

\begin{figure}[h]
  \centering
  \includegraphics[width=\textwidth]{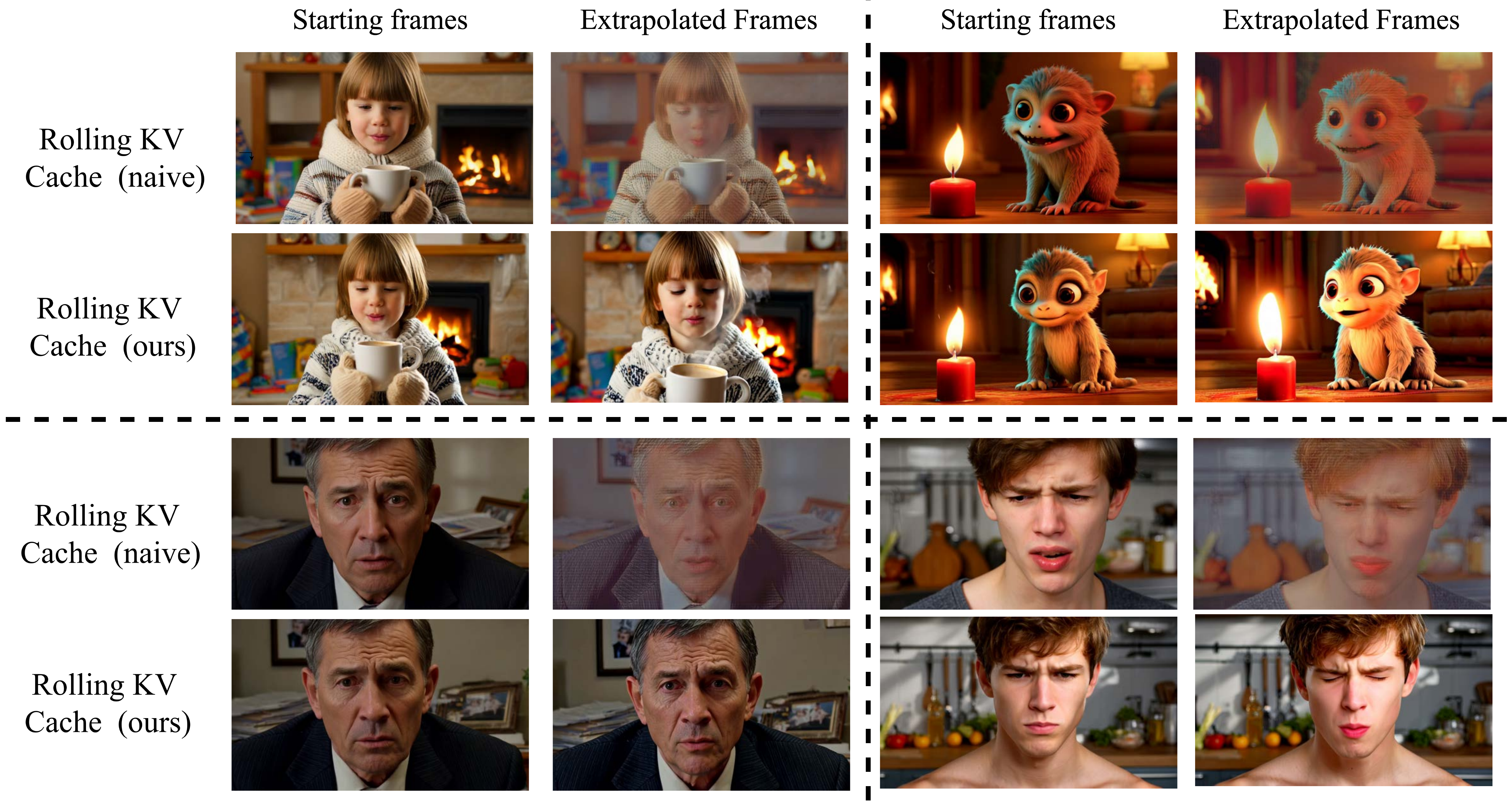}
  \caption{\textbf{Qualitative comparisons on video extrapolation.} We present a visual comparison between the naive baseline and our proposed technique for rolling KV cache-based video extrapolation. Compared to our method using local attention window training, extrapolated video frames from the naive baseline exhibit severe visual artifacts.}
  \label{fig:comp_10s}
\end{figure}

We qualitatively ablate two training settings for video extrapolation using the rolling KV cache technique. In the naive baseline, the model is trained such that every chunk always attends to the first chunk during denoising. In contrast, our proposed method restricts the attention window to prevent the model from attending to the first chunk when denoising the last chunk. As shown in Fig.~\ref{fig:comp_10s}, the naive baseline exhibits visual artifacts when extrapolating videos beyond the training context length, whereas our proposed solution mitigates this issue.

\section{VBench Scores Across All Dimensions}

\label{app:vbench}
\begin{wrapfigure}{r}{0.6\textwidth}
  \centering
  \includegraphics[width=\linewidth]{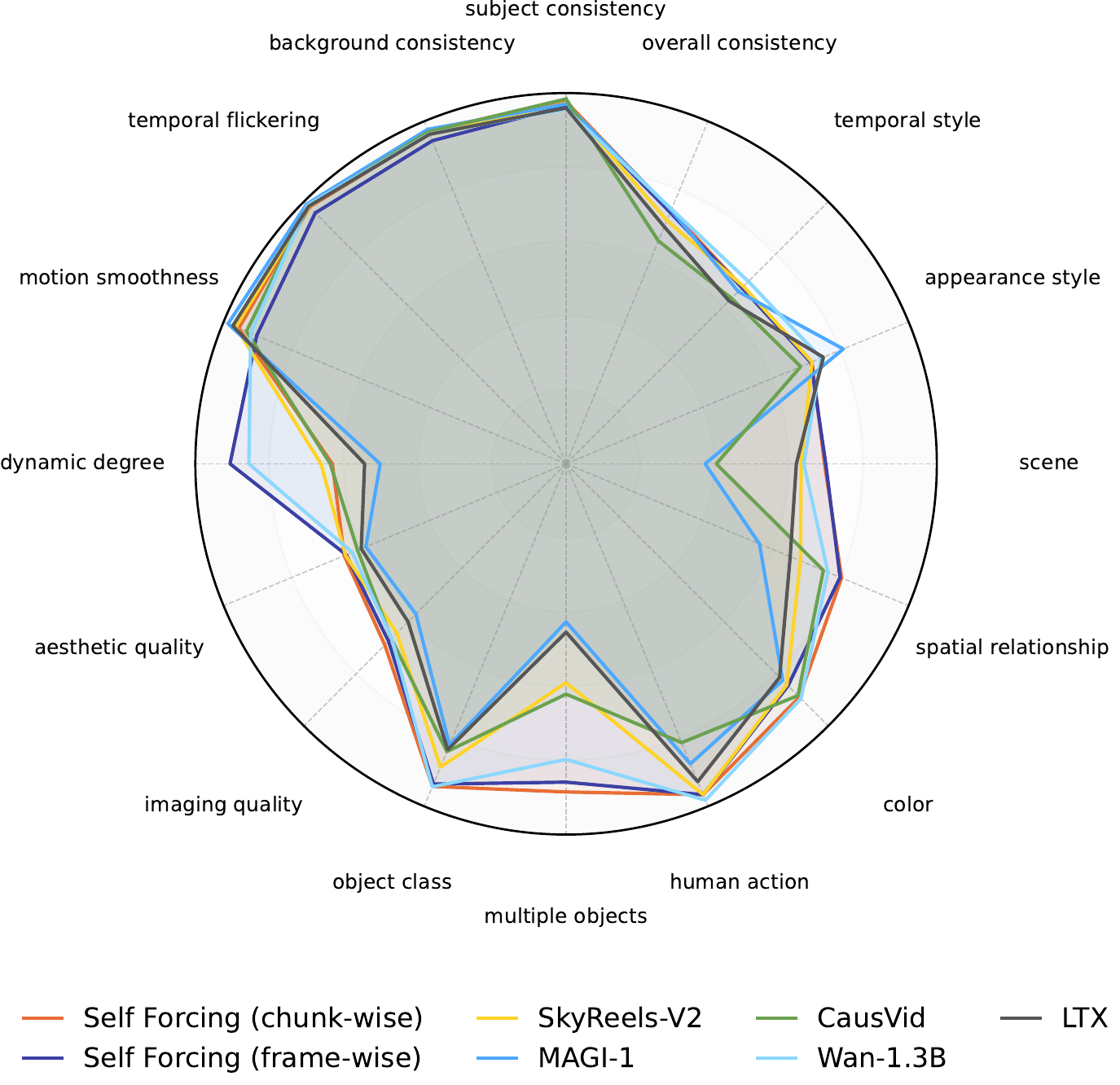}
  \caption{\textbf{VBench scores visualization.} We compare Self Forcing with SkyReels-V2~\cite{chen2025skyreels}, Wan2.1-1.3B~\cite{wang2025wan}, MAGI-1~\cite{magi1}, and CausVid~\cite{yin2025causvid} using all 16 VBench metrics.}
  \label{fig:radar}
  \vspace{-3em}
\end{wrapfigure}

In Fig.~\ref{fig:radar}, we evaluate Self Forcing~(both chunk-wise and frame-wise AR versions) using all 16 VBench metrics against representative models.
Self Forcing generally outperforms other models in terms of semantic alignment, evidenced by the high scores in scene, object class, multiple objects, and human action dimensions. Our methods also achieve good frame-wise quality, as indicated by the high scores in aesthetic quality and imaging quality.
Our frame-wise AR variant exhibits more dynamic motion~(high dynamic degree score) but worse temporal consistency~(worse background consistency, motion smoothness, and larger temporal flickering) than the chunk-wise AR variant.

\section{Broader Societal Impact}
\label{app:broader_impact}
Generative modeling—particularly for videos—carries significant potential for misuse. It can lead to serious societal consequences, most notably the spread of disinformation through deepfakes that become increasingly difficult to distinguish from authentic content. Additionally, these models can reinforce harmful stereotypes and amplify existing societal biases without careful governance and responsible deployment.

Our research on real-time video generation creates additional complexities, as it removes one of the practical barriers (computational cost) that currently limits widespread misuse. While our methods enable positive applications like creative content production and accessibility tools, we acknowledge the dual-use nature of this technology and encourage continued research into detection methods, watermarking techniques, and policy frameworks that can help mitigate potential harms.

\section{User Study Details}
\label{app:user_study}
\begin{figure}[h!]
  \centering
  \includegraphics[width=\textwidth]{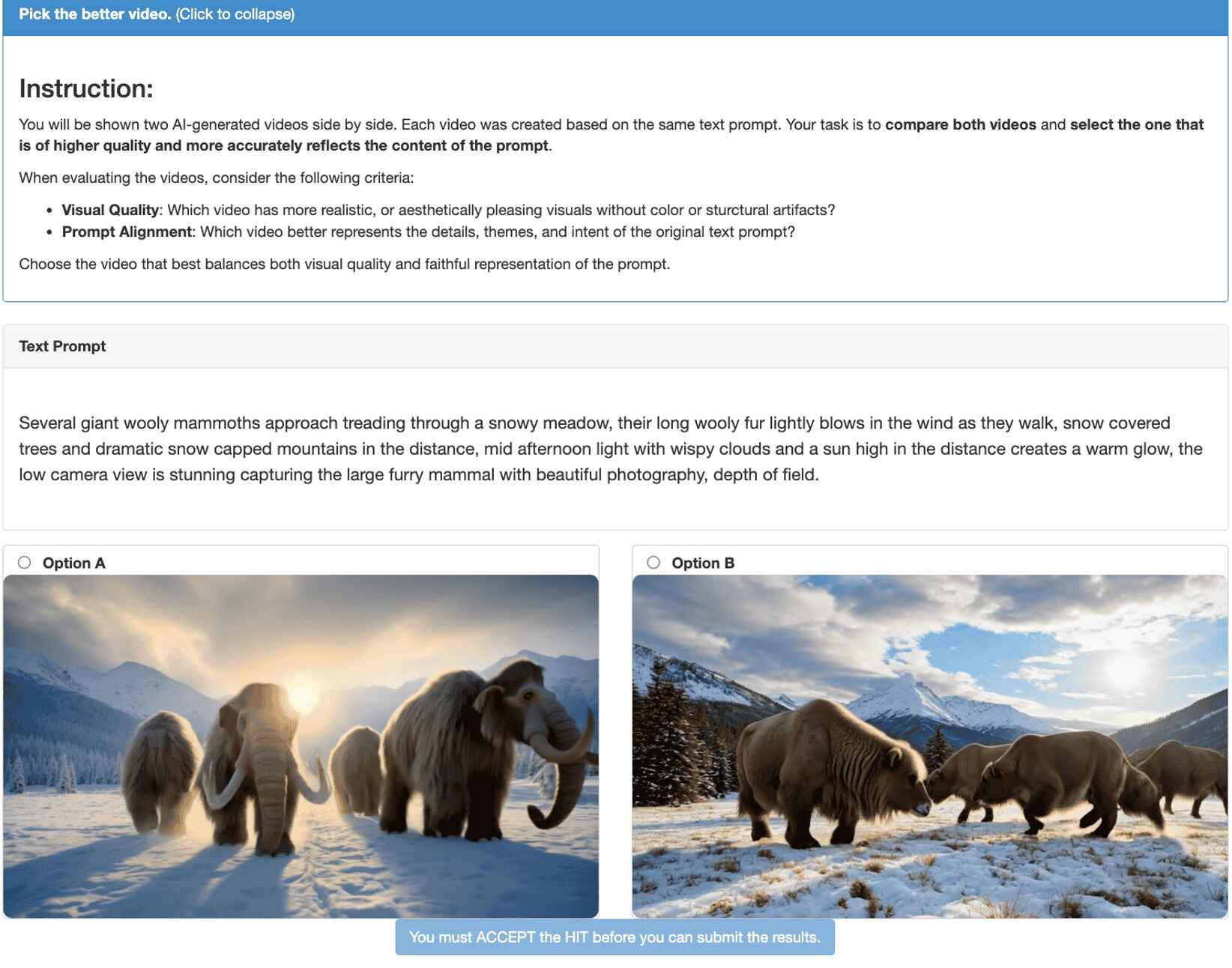}
  \caption{\textbf{User study instruction screenshots.}}
  \label{fig:user_study_instruction}
\end{figure}

In the user preference study, we show users two videos side by side using the same text prompt.
We ask the users to select the one that is overall better, considering both quality and prompt alignment. Detailed instructions are shown in Fig.~\ref{fig:user_study_instruction}.
We use all 1003 prompts from MovieGenBench~\cite{polyak2024movie} and each prompt is evaluated by a single user.

\end{document}